%% file: main.tex
\PassOptionsToPackage{table,xcdraw,dvipsnames}{xcolor}
\documentclass[10pt,twocolumn,letterpaper]{article}

 \usepackage{iccv}              % To produce the CAMERA-READY version

\input{preamble}

\input{command}

\def\paperID{901} % *** Enter the Paper ID here
\def\confName{ICCV}
\def\confYear{2025}

\title{Enhancing Partially Relevant Video Retrieval with Hyperbolic Learning}

\author{
Jun Li$^{1}\thanks{These authors contributed equally to this work.}$ , \ 
Jinpeng Wang$^{2*}\thanks{Corresponding author.}$ , \ 
Chaolei Tan$^{4}$,\
Niu Lian$^{1}$, \
Long Chen$^{4}$,\\
Yaowei Wang$^{3}$, \ 
Min Zhang$^{1}$, \ 
Shu-Tao Xia$^{2,3}$, \ 
Bin Chen$^{1}$ \\
{\small $^1$Harbin Institute of Technology, Shenzhen} \\
{\small $^2$Tsinghua Shenzhen International Graduate School, Tsinghua University} \\
{\small $^3$Research Center of Artificial Intelligence, Peng Cheng Laboratory} \\
{\small $^4$The Hong Kong University of Science and Technology} \\
{\small \texttt{220110924@stu.hit.edu.cn}\quad\Letter\ \texttt{wjp20@mails.tsinghua.edu.cn}}
}

\begin{document}
\maketitle
\input{sections/Abstract}    
\input{sections/Introduction}
\input{sections/RelatedWorks}
\input{sections/Method}

\input{sections/Experiments}
\input{sections/Conclusions}
{
    \small
    \bibliographystyle{ieeenat_fullname}
    \bibliography{main}
}

\clearpage
\appendix
% \input{appendix_sections/Method}
% \input{appendix_sections/Experiments}
\end{document}

\end{document}

%% file: preamble.tex
\usepackage{graphicx}
\usepackage{amsmath}
\usepackage{amssymb}
\usepackage{booktabs}
\usepackage{marvosym}
\definecolor{cvprblue}{rgb}{0.21,0.49,0.74}
\usepackage[pagebackref,breaklinks,colorlinks,allcolors=cvprblue]{hyperref}

\usepackage{url}            % simple URL typesetting
\usepackage{amsfonts}       % blackboard math symbols
\usepackage{nicefrac}       % compact symbols for 1/2, etc.
\usepackage{microtype}      % microtypography

\usepackage{makecell}
\usepackage{amsthm}
\usepackage{bm}
\usepackage{multicol}
\usepackage{colortbl}
\usepackage{multirow}
\usepackage{enumitem}
\usepackage{mathtools}
\usepackage{color}

\usepackage{algorithm}
\usepackage{listings}
\usepackage{natbib}
\usepackage{dsfont}

\usepackage{etoolbox}
\makeatletter
\AfterEndEnvironment{algorithm}{\let\@algcomment\relax}
\AtEndEnvironment{algorithm}{\kern2pt\hrule\relax\vskip3pt\@algcomment}
\let\@algcomment\relax
\newcommand\algcomment[1]{\def\@algcomment{\footnotesize#1}}
\renewcommand\fs@ruled{\def\@fs@cfont{\bfseries}\let\@fs@capt\floatc@ruled
  \def\@fs@pre{\hrule height.8pt depth0pt \kern2pt}%
  \def\@fs@post{}%
  \def\@fs@mid{\kern2pt\hrule\kern2pt}%
  \let\@fs@iftopcapt\iftrue}
\makeatother

%% file: command.tex
\definecolor{ForestGreen}{RGB}{34,139,34}

\renewcommand{\paragraph}[1]{\medskip\noindent\textbf{#1.~}}

\newcommand{\bma}{\bm{a}}

\newcommand{\bmc}{\bm{c}}

\newcommand{\bmf}{\bm{f}}

\newcommand{\bmq}{\bm{q}}

\newcommand{\bmt}{\bm{t}}

\newcommand{\bmv}{\bm{v}}
\newcommand{\bmw}{\bm{w}}
\newcommand{\bmx}{\bm{x}}
\newcommand{\bmy}{\bm{y}}
\newcommand{\bmz}{\bm{z}}

\newcommand{\bmF}{\bm{F}}

\newcommand{\bmQ}{\bm{Q}}

\newcommand{\bmV}{\bm{V}}

\newcommand{\modelname}{HLFormer}
\newcounter{qcounter}

\newcommand{\Lc}{\mathcal{L}}

\newcommand{\slorentz}{\mathbb{L}^n}
\newcommand{\linner}[2]{\langle {\bm{#1}},{\bm{#2}}\rangle_\Lc}
\newcommand{\lnorm}[1]{\lVert #1\rVert_\Lc}

\newcommand{\stangent}[1]{\mathcal{T}_{\bm{#1}}\mathbb{L}^n}

\newcommand{\slogmap}[2]{\log_{\bm{#1}}(\bm{#2})}
\newcommand{\sexpmap}[2]{\exp_{\bm{#1}} (\bm{#2})}
\newcommand{\riemanntensor}[1][x]{\mathfrak{g}_\bm{x}^K}

\newcommand{\Real}{\mathbb{R}}
\newcommand{\sriemanntensor}[1][x]{\mathfrak{g}_{\bm{x}}}

%% file: sections/Abstract.tex
\begin{abstract}
Partially Relevant Video Retrieval (PRVR) addresses the critical challenge of matching untrimmed videos with text queries describing only partial content. Existing methods suffer from geometric distortion in Euclidean space that sometimes misrepresents the intrinsic hierarchical structure of videos and overlooks certain hierarchical semantics, ultimately leading to suboptimal temporal modeling. To address this issue, we propose the   first hyperbolic modeling framework for PRVR,  namely HLFormer, which leverages  hyperbolic space learning to compensate for the suboptimal hierarchical modeling capabilities of Euclidean space. Specifically, HLFormer integrates the Lorentz Attention Block  and Euclidean Attention Block  to encode video embeddings in hybrid spaces, using the Mean-Guided Adaptive Interaction Module to dynamically fuse features. Additionally, we introduce a Partial Order Preservation Loss to enforce ``$\text{text} \prec \text{video}$'' hierarchy through Lorentzian cone constraints. This approach further enhances cross-modal matching by reinforcing partial relevance between video content and text queries. Extensive experiments show that HLFormer outperforms state-of-the-art methods. Code is released at \href{https://github.com/lijun2005/ICCV25-HLFormer}{https://github.com/lijun2005/ICCV25-HLFormer}.
\end{abstract}

%% file: sections/Introduction.tex
\section{Introduction}
\label{sec:intro}
\begin{figure}[t]
    \centering
    \begin{subfigure}[t]{0.45\textwidth}
        % \centering
        \includegraphics[width=\textwidth]{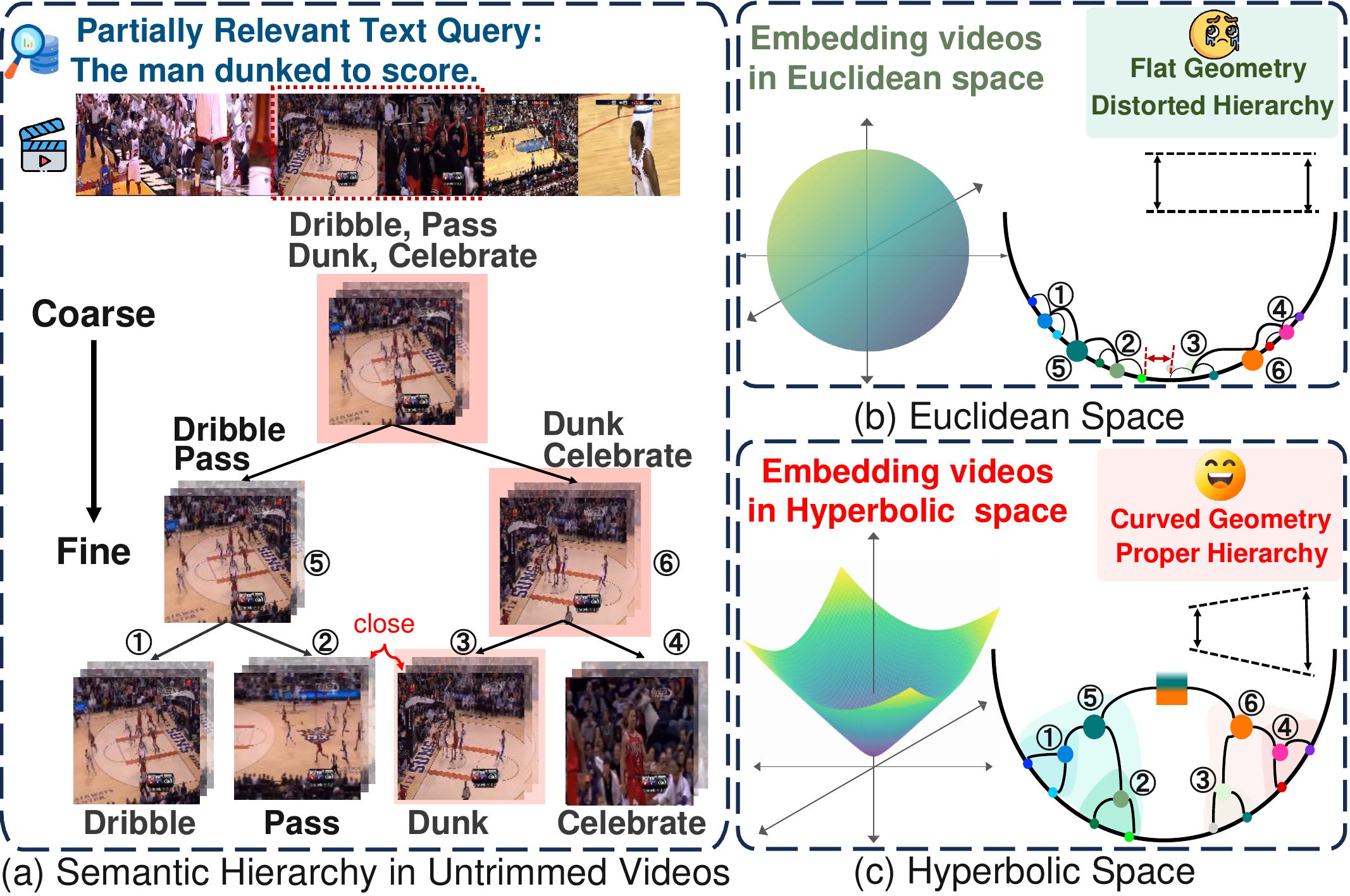}
    \end{subfigure}
    \caption{(a) Modeling the semantic hierarchy in untrimmed videos helps Partially Relevant Video Retrieval (PRVR).
    (b) Euclidean space is less effective in modeling semantic hierarchy due to the flat geometry. Data points with distant hierarchical relation may be close. 
    (c) Hyperbolic space allows larger cardinals when approaching the edge, which is preferable to preserve the hierarchy. }
    \label{fig:Intro}
\end{figure}
Text-to-video retrieval (T2VR) \cite{dong2018predicting, chen2020fine, miech2019howto100m, liu2019use, li2019w2vv++, faghri2017vse++, dong2019dual, dong2021dual, dong2022reading} is a fundamental module in many search applications and a popular topic in multi-modal learning. 
While most T2VR models are developed for short clips or pre-trimmed video segments, they may face challenges where user queries describe only \emph{partial} content in the video. 
This practical issue in real-world usage promotes a more challenging setting of \emph{partially relevant video retrieval} (PRVR) \cite{ms-sl}, which aims to match each text query with the best \emph{untrimmed} video. 

Due to unlabeled moment timestamps, PRVR requires solid abilities on 
(\textbf{i}) identifying key moments in videos for extracting informative features and 
(\textbf{ii}) learning robust cross-modal representations to match text queries and videos precisely. 
Prior arts have developed preliminary solutions on both aspects, while challenges remain. 
For (i), MS-SL \cite{ms-sl} exhaustively enumerated consecutive frame combinations through multi-scale sliding windows, which inevitably engaged redundancy, noise, and a high computational complexity in extracting moment features. 
GMMFormer \cite{GMMFORMER,gmmformerv2} improved efficiency by leveraging Gaussian neighborhood priors to traverse each timestamp and discover potential key moments. 
However, it may still be hard to distinguish adjacent or semantically similar candidate moments. 
Though DL-DKD \cite{DKD} neatly benefited from the pretrained CLIP \cite{CLIP} to enhance text-frame alignment, the temporal generalizability is bounded by the text-\emph{image} teacher model. 
For (ii), most existing solutions inherited similar ideas from classic T2VR, \eg, ranking and contrastive learning, at a holistic level, but important characteristics of PRVR, \eg, partial relevance and semantic entailment, are still under-explored. 

In this paper, we take a hierarchical perspective to review the task, in the belief that videos naturally exhibit semantic hierarchy. 
As illustrated in \cref{fig:Intro}(a), an untrimmed video can be regarded as a progression from frames to informative segments (\eg, Dunk), extended moments, and ultimately, the whole. 
Leveraging this intrinsic property is expected to benefit long video understanding. 
In particular for PRVR, the hierarchical prior provides positive guidance to arrange the moment features. 
Meanwhile, the supervisory signals from query-video matching can activate moment extraction more precisely through \emph{implicit} bottom-up modeling. 
Exploring hierarchical features is never trivial. 
Unfortunately, existing PRVR approaches relying on Euclidean space are less effective in modeling the desired patterns in the flat geometry. 
We present \cref{fig:Intro}(b) to exemplify this: two embeddings with a distant hierarchical relation may be spatially close to each other, as marked by the red arrows. 
Biased representation will increase the difficulty in disentangling informative moments from background, which limits the robustness in cross-modal matching considering partial relevance. 

Inspired by the emerging success of hyperbolic learning \cite{poincareembedding, dsrl, lorentizan, meru, hyperbolicvisiontransformer}, which takes advantage of exponentially expanding metric in non-Euclidean space to better capture hierarchical structure (\cref{fig:Intro}(c)), we introduce \modelname{}, a sincere exploration of hyperbolic learning to enhance PRVR. 
On \textbf{\emph{temporal modeling}}, we carefully design a dual-branch strategy to capture informative moment features comprehensively. 
Specifically, for the hyperbolic branch, we develop a Lorentz Attention Block (LAB) with the hyperbolic self-attention mechanism. 
With the implicit hierarchical prior through end-to-end matching optimization, LAB learns to activate informative moment features relevant to queries and distinguish them from noisy background in the hyperbolic space, compensating for the limitations of Euclidean attention in capturing hierarchical semantics. 
We integrate dual-branch moment features with a Mean-Guided Adaptive Interaction Module (MAIM), which is lightweight but effective. 
On \textbf{\emph{cross-modal matching}}, drawing on the intrinsic 
``$\text{text} \prec \text{video}$'' hierarchy in PRVR where textual queries are subordinate to their paired videos, we introduce a Partial Order Preservation (POP) loss that geometrically confines text embeddings within hyperbolic cone anchored by corresponding video representations in an auxiliary Lorentzian manifold. This hierarchical metric alignment ensures semantic consistency between localized text semantics and their parent video structure while preserving partial relevance.

Empirical evaluations on three benchmark datasets: ActivityNet Captions \cite{krishna2017dense}, Charades-STA \cite{gao2017tall}, and TVR \cite{lei2020tvr} establish \modelname{}'s state-of-the-art performance. Ablation studies confirm the necessity of hyperbolic geometry for hierarchical representation and the critical role of explicitly relational constraints in  Partial Order Preservation Loss. Meanwhile, visual evidences further reveal  that hyperbolic learning can enhance discriminative representation while maintaining video-text entailment, sharpening moment distinction and improving query alignment.

The primary contributions can be summarized as follows: 
\begin{itemize}
    \item[$\bullet$] We propose to enhance PRVR with hyperbolic learning, including a Lorentz attention block with hierarchical priors to enhance the moment feature extraction, which collaborates with Euclidean attention and hybrid-space fusion.
    \item[$\bullet$]  We design a partial order preservation loss that geometrically enforces the ``\text{text} $\prec$ \text{video}'' hierarchy through hyperbolic cone constraints, strengthening partial relevance.
    \item[$\bullet$]  Extensive experiments on three benchmarks validate \modelname{}'s superiority, with analyses confirming the efficacy of hyperbolic modeling and geometric constraints.
\end{itemize}

%% file: sections/RelatedWorks.tex
\section{Related Works}
\label{sec:related_works}
\subsection{Partially Relevant Video Retrieval}
With the growth of video content \cite{liu2025protecting,fang2025grounding,wang2025empirical}, video retrieval has become a key research area. Given a text query, Text-to-Video Retrieval (T2VR) \cite{dong2018predicting, chen2020fine, miech2019howto100m, liu2019use, li2019w2vv++, faghri2017vse++, dong2022reading,wang2022hybrid,liu2022multi,wang2024hugs}  focuses on retrieving fully relevant videos from pre-trimmed short clips. Video Corpus Moment Retrieval (VCMR) \cite{song2021spatial, lei2020tvr,jsg,tan2024siamese} aims to localize specific moments within videos from a large corpus. Partially Relevant Video Retrieval (PRVR) \cite{ms-sl,DKD,GMMFORMER,gmmformerv2,bgmnet,PEAN,t-d3n,cho2025ambiguity}, a more recent task introduced by  \citet{ms-sl}, aims to retrieve partially relevant videos from large, untrimmed long video collections. Unlike T2VR, PRVR must address the challenge of partial relevance, where the query pertains to only a specific moment of the video. Though the first stage of VCMR is similar  to  PRVR, VCMR requires moment-level annotations, limiting  scalability.

Existing methods enhance PRVR retrieval from various perspectives. MS-SL \cite{ms-sl} defines the PRVR task as a Multi-instance Learning, providing a strong baseline with explicit redundant clip embeddings. GMMFormer \cite{GMMFORMER,gmmformerv2} and PEAN \cite{PEAN} propose implicit clip modeling to improve efficiency. DL-DKD \cite{DKD} achieves great results through dynamic distillation of CLIP \cite{CLIP}. BGM-Net \cite{bgmnet} exploits an instance-level matching scheme for pairing queries and videos. 
However, these methods predominantly rely on Euclidean space, which sometimes distort the hierarchical structures in untrimmed long videos. Consequently, they fail to fully exploit video hierarchy priors.
To overcome this issue, we propose \modelname{} to enhances PRVR by implicitly capturing hierarchical structures through hyperbolic learning.

\subsection{Hyperbolic Learning}
Hyperbolic learning has attracted significant attention for its effectiveness in modeling hierarchical structures in real-world datasets.
Early studies in computer vision tasks explored hyperbolic image embeddings from image-label pairs \cite{poincareembedding,Khrulkov_2020_CVPR}, while subsequent progress extended hyperbolic optimization to multi-modal learning. MERU \cite{meru} and HyCoCLIP \cite{hycoclip} notably surpassed Euclidean counterparts like CLIP \cite{CLIP} via hyperbolic space adaptation. Applications span semantic segmentation \cite{segmenation1,segmentation2}, recognition tasks (skin \cite{skin}, action \cite{long2020searching}), meta-learning \cite{hyperbolicvisiontransformer}, and detection frameworks (violence \cite{hypvd,dsrl}, anomalies \cite{anomalydetection}). Recent advances in fully hyperbolic neural networks \cite{net1,net2,net3,net4,net5} further underscore their potential.
Motivated by them, we present the first study to explore the potential of hyperbolic learning for PRVR. Unlike other methods such as DSRL \cite{dsrl} and HOVER \cite{shi2024hover}, our approach utilizes hyperbolic space to compensate for the limitations of Euclidean space in capturing the hierarchical structure of untrimmed long videos. Furthermore, we introduce the Partial Order Preservation Loss to explicitly capture the partial relevance between video and text in hyperbolic space,  improving retrieval performance.

%% file: sections/Method.tex
\section{Method}
\begin{figure*}[t]
    \centering
    \includegraphics[width=\textwidth]{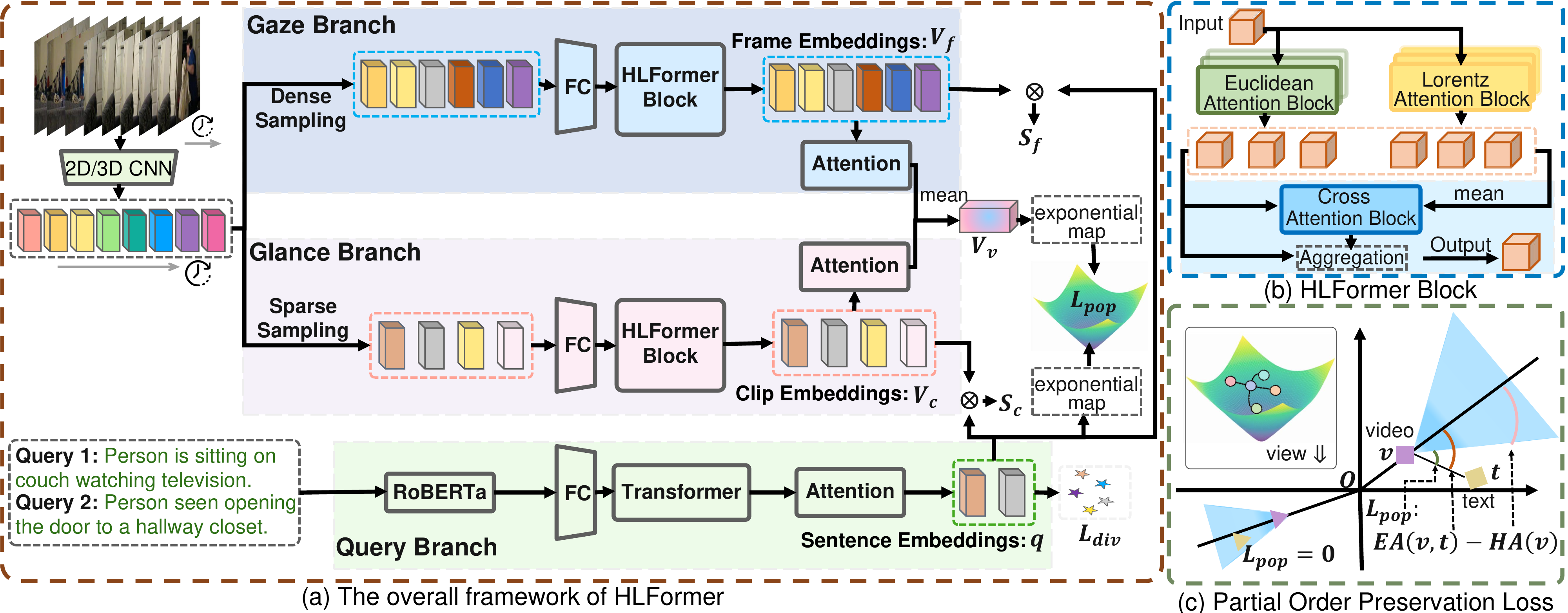}
    \caption{Overview of \modelname{}. (a) The sentence embedding $\bmq$ is obtained via the query branch, while the gaze and glance branches encode the video, producing frame-level embedding $\bmV_f$ and clip-level embedding $\bmV_c$ and forming the video representation $\bmV_v$. $\bmq$ learns query diversity through $L_{div}$ and computes similarity scores $S_f$ and $S_c$, while preserving partial order relations with $\bmV_v$ using $L_{pop}$. (b) \modelname{} block combines parallel Lorentz and Euclidean attention blocks for  multi-space encoding, with a Mean Guided Adaptive Interaction Module for dynamic aggregation. (c) Partial Order Preservation Loss  ensures the text query embedding $\bmt$ lies within the cone defined by the video embedding $\bmv$. 
    The loss is zero if $t$ is inside the cone.}
    \label{fig:arc}
\end{figure*}
\subsection{Preliminaries}
\noindent \textbf{Hyperbolic Space}  \quad Hyperbolic spaces are Riemannian manifolds with a constant negative curvature $K$, 
 contrasting with the zero-curvature (flat) geometry of Euclidean spaces. 
Among several isometrically equivalent hyperbolic models, we adopt the Lorentz model \cite{lorentzraw} for its numerical stability and computational efficiency, with $K$ set to -1 by default.

\noindent \textbf{Lorentz Model}  \quad 
Formally, an $n$-dimensional Lorentz model is the Riemannian manifold $\slorentz = (\Lc^n, \sriemanntensor)$. $\sriemanntensor=\mathop{\mathgroup\symoperators diag}(-1,1,\cdots,1)$ is the Riemannian metric tensor. 
Each point in $\slorentz$ has the form  $\bmx = \left[ x_{0},\bmx_{s} \right] \in\Real^{n+1}, x_{0} = \sqrt{||\bmx_{s}||^2+1} \in \Real$. Following \citet{net2}, we denote  $x_0$ as \textit{time axis} and  $\bmx_{s}$ as \textit{spatial axes}. $\Lc^n$ is given by:
\begin{equation}
\Lc^n \coloneqq \{\bm{x}\in \Real^{n+1}  \mid \linner{x}{x}=-1, x_{0} >0\},
\end{equation}
and the Lorentzian inner product given by:
\begin{equation}
\linner{x}{y}\coloneqq-x_{0} y_{0}+\bm{x}_{s}^{\top}\bm{y}_{s}.
\label{eq:inner}
\end{equation}
Here $\Lc^n$ is the upper sheet of hyperboloid in a $(n+1)$ dimensional Minkowski space with the origin $\bm{o} = (1, 0, \cdots, 0)$. 

\noindent \textbf{Tangent Space}  \quad 
The tangent space at  $\bm{x}\in \slorentz$ is a Euclidean space  that is orthogonal to it, defined as:
\begin{equation}
\stangent{x}\coloneqq\{\bm{y}\in \Real^{n+1} \mid \linner{y}{x}=0\}.
\label{eq:tangent}
\end{equation}
Where $\stangent{x}$ is a Euclidean subspace of $\Real^{n+1}$. In particular, the tangent space at the origin $\bm{o}$ is denoted as $\stangent{o}$. 

\noindent \textbf{Logarithmic and Exponential Maps}  \quad  The mutual mapping between the hyperbolic space $\slorentz$ and the Euclidean subspace $\stangent{x}$ can be realized by logarithmic and exponential maps.
The exponential map $\exp_{\bm{x}}(\bm{z})$
% $:\stangent{x}\rightarrow\slorentz$ 
can map any tangent vector $\bm{z} \in \stangent{x}$ to $\slorentz$, written as:
\begin{equation}\label{eq:exp}
\sexpmap{x}{z} = \cosh(\lVert{\bmz}\rVert_\Lc){\bmx} + \sinh(\lVert{\bmz}\rVert_\Lc)\frac{{\bmz}}{\lVert{\bmz}\rVert_\Lc}, 
\end{equation}
where $ \lVert{\bmz}\rVert_\Lc = \sqrt{\linner{z}{z}}$ and the logarithmic map $\slogmap{{x}}{{y}}$
% $:\slorentz \rightarrow \stangent{x}$
plays an opposite role to map $\bm{y} \in \slorentz$ to $\stangent{x}$ as follows:
\begin{equation}
\slogmap{{x}}{{y}} = \frac{\operatorname{arcosh}(-\linner{x}{y})}{\sqrt{(-\linner{x}{y})^2-1}}({\bmy}+(\linner{x}{y}){\bmx}).
\label{eq:log}
\end{equation}
\noindent \textbf{Lorentzian centroid } The weighted centroid with respect to the squared Lorentzian distance, which solves $\min_{\mu \in \mathbb{L}^n}\sum_{i=1}^{m}\nu_i d^2_{\mathcal{L}}( {\bm{x}_i},  {\mu})$, with $ {\bm{x}_i} \in \mathbb{L}^n$ and $\nu_i\geq0, \sum_{i=1}^{m}\nu_i > 0$, is denoted as:
\begin{equation}\label{eq:centroid}
	\mu = \frac{\sum_{i=1}^{m}\nu_i \bm{x}_i}{\left|||\sum_{i=1}^{m}\nu_i {\bm{x}_i}||_{\mathcal{L}}\right|}.
\end{equation}

\subsection{Problem Formulation and Overview}
\label{sec:overview}
Partially Relevant Video Retrieval (PRVR) aims to retrieve videos containing a moment semantically
relevant to a given text query, from a large corpus of untrimmed videos. In the PRVR database, each video has multiple moments and is associated with multiple text descriptions, with each text description corresponding to a specific moment of the related video. Critically, the temporal boundaries of these moments (\textit{i.e.},  start and end time points) are not annotated.  

In this paper, we introduce \modelname{}, the first hyperbolic modeling approach designed for PRVR.  The proposed framework encompasses three key components: text query  representation encoding, video representation encoding, and similarity computation, as illustrated in \cref{fig:arc} (a).

\noindent \textbf{Text Representation}  \quad 
Given a text query of $ N_q $ words, we first use a pre-trained RoBERTa \cite{liu2019roberta} model to extract word-level features, which are then projected into a lower-dimensional space via a fully connected (FC) layer. A standard Transformer \cite{vaswani2017attention} layer is applied to obtain a sequence of $ d $-dimensional contextualized  feature vectors, $ \bmQ = \{\bmq_i\}_{i=1}^{N_q} \in \mathbb{R}^{N_q\times d} $. 
Finally, we utilize a simple attention mechanism to get the sentence embedding $ \bmq \in \mathbb{R}^d $:
\begin{equation}
\bmq = \sum_{i=1}^{N_q} \bma_i^q \times \bmq_i, \quad \bma^q = \text{softmax}(\bmw \bm{Q^{\top}}),    
\label{eq:simpleattention}
\end{equation}
where $ \bmw \in \mathbb{R}^{1\times d} $ is a trainable vector, and $ \bma^q \in \mathbb{R}^{1\times N_q} $ represents the attention vector.

\noindent \textbf{Video Representation} \quad 
Given an untrimmed video, we first extract embedding features using a pre-trained 2D or 3D CNN. Then we utilize the gaze branch and glance branch to capture frame-level and clip-level multi-granularity video representations, respectively. In the gaze branch, we densely sample $M_f$ frames, denoted as $\bmF \in \mathbb{R}^{M_f \times D}$, where $D$ is the frame feature dimension. The sampled frames are processed through a fully connected (FC) layer to reduce the dimensionality to $d$, followed by the \textbf{\modelname{} block} to obtain frame embeddings $\bm{V_f} = \{\bmf_i\}_{i=1}^{M_f} \in \mathbb{R}^{M_f \times d}$, capturing semantically rich frame-level information for fine-grained relevance assessment to the query. 
The glance branch down-samples the input along the temporal dimension to aggregate frames into clips. Following MS-SL \cite{ms-sl}, a fixed number $ M_c $ of clips is sparsely sampled by mean pooling over consecutive frames. A fully connected layer is applied to the pooled clip features, followed by the \textbf{\modelname{} block}, generating clip embeddings $ \bm{V_c} = \{\bmc_i\}_{i=1}^{M_c} \in \mathbb{R}^{M_c \times d} $. These embeddings capture adaptive clip-level information, enabling the model to perceive relevant moments at a coarser granularity.

\noindent \textbf{Similarity Computation} \quad
To compute the similarity between a text-video pair $(\mathcal{T}, \mathcal{V})$, we first measure the above- mentioned embeddings $\bmq$, $\bm{V_f}$ and $\bm{V_c}$.  Then, we employ cosine similarity along with a max operation to calculate the frame-level and clip-level similarity scores:
\begin{equation}
\begin{aligned}
S_f(\mathcal{T}, \mathcal{V}) = \text{max} \{ \text{cos}(\bmq, \bmf_1),..., \text{cos}(\bmq, \bmf_{M_f})\}, \\
S_c(\mathcal{T}, \mathcal{V}) = \text{max} \{\text{cos}(\bmq, \bmc_1),..., \text{cos}(\bmq, \bmc_{M_c})\}.
\label{eq:simscore}
\end{aligned}
\end{equation}
Next, we compute the overall text-video pair similarity:
\begin{equation}\label{eq:sim}
S(\mathcal{T}, \mathcal{V}) = \alpha_f S_f(\mathcal{T}, \mathcal{V}) + \alpha_c S_c(\mathcal{T}, \mathcal{V}),
\end{equation}
where $ \alpha_f, \alpha_c \in [0, 1] $ are hyper-parameters satisfying  $ \alpha_f + \alpha_c = 1 $. Finally,  we retrieve and rank partially relevant videos based on the computed similarity scores.

\subsection{\modelname{} Block}
The \modelname{} Block constitutes the core of our method. As shown in \cref{fig:arc} (b), it comprises three key modules: (\textbf{i}) Euclidean Attention Block, capturing fine-grained visual features in Euclidean space; (\textbf{ii}) Lorentz Attention Block,  projecting video embeddings into hyperbolic Lorentz space for  capturing the hierarchical structures of video; (\textbf{iii}) Mean-Guided Adaptive Interaction Module, dynamically fusing hybrid-space features. We describe the details below.

\noindent \textbf{Euclidean Attention Block} \quad
Given M feature embeddings $ \bmx \in \mathbb{R}^{M \times d} $, where $ d $ is the feature dimension, the Euclidean Attention Block utilizes Euclidean Gaussian Attention \cite{GMMFORMER} to capture multi-scale visual features, expressed as:
\begin{equation}\label{eq:euclidean-attention}
    \text{GA}(\bmx) = \text{softmax}\left(\mathcal{M}_{\sigma}^{g} \odot \frac{ \bmx W^q ( \bmx W^k)^{\top}}{\sqrt{d_h}}\right) \bmx W^v,
\end{equation}
where $\mathcal{M}_{\sigma}^{g}$ is the Gaussian matrix with elements $\mathcal{M}_{\sigma}^{g}(i, j) = \frac{1}{2\pi} e^{-\frac{(j-i)^2}{\sigma^2}}$, and $\sigma^2$ denotes the variance. By varying $\sigma$, feature interactions at different scales are modeled, generating video features with multiple receptive fields. $ W^q, W^k, W^v $ are linear projections, 
% of $ x $ into query, key, and value, respectively,
while $ d_h $ is the latent attention dimension, $ \odot $ denotes element-wise product. Finally, We replace the self-attention in Transformer block with Euclidean Gaussian  attention to form the Euclidean Attention Block.

\noindent \textbf{Lorentz Attention Block} \quad  
Given extracted Euclidean video embeddings \( \bmx^E_{\text{in}} \in \mathbb{R}^{M \times d} \),
% , where \( M \) denotes the number of frames or clips and \( d \) the feature dimension, 
we first project it to \( \mathbb{R}^{M \times n} \) via a linear layer and apply scaling. Let \( \bm{o}:= [1, 0, \dots, 0] \) be the origin on the Lorentz manifold, satisfying \( \langle \bm{o}, [0, \bmx_{\text{in}}^E] \rangle_{\mathcal{L}} = 0 \).
Thus, \( [0, \bmx_{\text{in}}^E] \) can be interpreted as a vector in the tangent space at \( \bm{o} \). The Lorentz embedding is then obtained via the exponential map \cref{eq:exp}:
\begin{equation}\label{eq:in}
    \bm{\bm{x^{\mathcal{L} }_{\text{in}}}} = \mathrm {exp}_{\bm{o} }\left ( \left [0,\beta \bmx^{E}_{\text{in}}  W_1 \right ]  \right )  \in \slorentz, \mathbb{R}^{M \times (n+1)},
\end{equation}
where $W_1$ denotes the linear layer, $\beta$ is a  learnable scaling factor to prevent numerical overflow.
%%%%%%%%%%%%%%%%%%%%%%%%%%%

Having obtained the Lorentz embedding $\bm{x^{\mathcal{L} }_{\text{in}}}$, which inherently exhibits a prominent hierarchical structure due to the hyperbolic space properties, we next design a Lorentz linear transformation and Lorentz self-attention module to capture and fully leverage the hierarchical priors.

%%%%%%%%%%%%lorentz linear layer
Inspired by prior studies \cite{net2,net5}, we redefine the Lorentz linear layer to learn a matrix  $\bm{M} = \begin{bmatrix} \bm{p}^{\top} \\ \bm{W} \end{bmatrix} $, where  $\bm{p} \in \mathbb{R}^{n+1} $ is a weight parameter and  $\bm{W} \in \mathbb{R}^{m \times (n+1)} $ ensures that  $\forall \bm{x} \in \mathbb{L}^{n} $,  $f_{\bm{x}}(\bm{M})\bm{x} \in \mathbb{L}^{m} $. 
Specifically, the transformation matrix $f_{\bm{x}}(\bm{M})$ is expressed as:
\begin{equation}
    f_{\bm{x}}(\bm{M})=f_{\bm{x} }\left ( \begin{bmatrix}
 \bm{p}^{\top}  \\ \bm{W} 
\end{bmatrix} \right ) =\begin{bmatrix}
 \frac{\sqrt{\left \|  W\bm{x}  \right \|^{2}+1 }}{\bm{p}^{\top}\bm{x}  }\bm{p}^{\top} \\
\bm{W}.
\end{bmatrix}
\end{equation}
Adding other components including normalization, the final definition of the  Lorentz Linear layer becomes:
\begin{equation}
\bm{y} = \texttt{HL}(\bm{x}) = \left[\begin{smallmatrix}
\sqrt{\lVert \phi\left ( \bm{W}\bm{x},\bm{p}  \right ) \rVert^2 +1}  \\
\phi(\bm{W}\bm{x}, \bm{p})
\end{smallmatrix}\right],
\label{eq:linearlayer}
\end{equation}
with operation function:
\begin{equation}
    \phi \left ( \bm{W}\bm{x},\bm{p}  \right ) = \frac{\lambda\left (  \bm{p}^{\top}\bm{x}+  b^{\prime}   \right )  }{\left \| \bm{W}h\left ( \bm{x}  \right )+\bm{b}    \right \| }\left ( \bm{W}h\left ( \bm{x}  \right )+\bm{b} \right ),
\end{equation} 
where $\bm{b}$ and $b^{\prime}$ are bias terms, $\lambda > 0$ regulates the scaling range.  $h$ denotes the activation function.

Based on the Lorentz Linear Layer, we propose a Lorentz self-attention module that integrates Gaussian constraints into feature interactions, enabling multiscale and hierarchical video embeddings in hyperbolic space. Specifically, given a hyperbolic video embedding $\bm{x^{\mathcal{L} }_{\text{in}}} \in \slorentz, \mathbb{R}^{M \times (n+1)} $, we first obtain the attention query $\mathcal{Q} $, key  $\mathcal{K} $, and value  $\mathcal{V} $ using \cref{eq:linearlayer}, all in the shape of $\mathbb{R}^{M \times (n+1)}$.
We  calculate attention scores based on \cref{eq:centroid} and  apply a Gaussian matrix $\mathcal{M}^g_{\sigma} \in \mathbb{R}^{M \times M}$ for element-wise multiplication with the  score matrix to obtain a multi-scale receptive field. The  output is defined as $\bm{x^{\mathcal{L} }_{\text{out}}} = \{\bm{\mu}_1, \ldots, \bm{\mu}_{|\mathcal{Q}|}\} \in \mathbb{R}^{M \times (n+1)} $:
\begin{equation}
\begin{aligned}    S_{ij} &= \frac{\exp (\frac{-d_{\Lc}^2(\bm{q}_i, \bm{k}_j) \odot  \mathcal{M}^g_{\sigma}(i,j)}{\sqrt{(n+1)}})}{\sum_{k=1}^{|\mathcal{K}|} \exp (\frac{-d_{\Lc}^2(\bm{q}_i, \bm{k}_k) \odot  \mathcal{M}^g_{\sigma}(i,k)}{\sqrt{(n+1)}})}, \\
\bm{\mu}_i&=\frac{\sum_{j=1}^{|\mathcal{K}|} S_{ij}\bm{v}_{j}}{ \big | \lnorm{\sum_{k=1}^{|\mathcal{K} |}S_{ik}\bm{v}_k} \big|},
\end{aligned}
\end{equation}
the squared Lorentzian distance  $d^2_\mathcal{L}(\bm{a}, \bm{b})=-2-2\linner{a}{b}$.

After computing  $\bm {x^{\mathcal{L}}_{\text{out}}}  $, we apply the logarithmic map  \cref{eq:log}, while discarding the time axis, to obtain the Euclidean space embedding  $ \bmx^{{E}}_{\text{mid}}  $. Then, the output  $ \bmx^{{E}}_{\text{out}}  $ is obtained through a Linear Layer followed by rescaling:
\begin{equation}
\begin{aligned}
    \bmx^{E }_{\text{mid}}&= \texttt{drop\_time\_axis} (\slogmap{o}{{x^{\mathcal{L}}_{\text{out}}}}) \in \mathbb{R}^{M \times n},\\
    \bmx^{E }_{\text{out}} &= \frac{ \bmx^{E}_{\text{mid}} W_2 }{\beta} \in \mathbb{R}^{M \times d},
\end{aligned}
\end{equation}
where $W_2 \in \mathbb{R}^{n \times d }$, $\beta$ is the scale factor in \cref{eq:in}. 
Finally, We replace the self-attention in Transformer block with Lorentz  attention to form the Lorentz Attention Block.

\noindent \textbf{Mean-Guided Adaptive Interaction Module} \quad We arange $N_{\mathcal{L}}$ Lorentz and $N_E$ Euclidean Attention Blocks in parallel to construct $N_O$ Gaussian Attention Blocks for multi-scale hybrid-space video embeddings.  To integrate these features, we introduce a Mean-Guided Adaptive Interaction Module, which utilizes globally pooled features to compute dynamic aggregation weights. Specifically, we first obtain the global query $ \bm{\varphi} \in \mathbb{R}^{1 \times d} $ and compute aggregation weights via a Cross Attention Block consisting of  a cross-attention layer (CA) followed by a fully connected layer (FC):
\begin{equation}
    \begin{aligned}
            \bm{\varphi}& = \text{Mean}( \bmx_{\sigma_{1}}, \bmx_{\sigma_{2}},..,\bmx_{\sigma_{N_o}}), \\
    w_i &= \text{FC}(\text{CA}(\bm{\varphi}, \bmx_{\sigma_{i}}, \bmx_{\sigma_{i}})), i = 1, 2,..., N_o, \\
        \Tilde{w}_{i,j} &= \frac{e^{w_{i,j} / \tau}}{\sum_{k=1}^{N_o}e^{w_{k, j} / \tau}}, j = 1,..., M,\\
    \bm{\Tilde{x}_j} &=  \sum_{i=1}^{N_o}  \Tilde{w}_{i,j} \bmx_{\sigma_{i},j}, j = 1,..., M, \\
        \bmx_{\text{MAIM}} &= \text{Concat}(\bm{\Tilde{x}}_1, \bm{\Tilde{x}}_2,..., \bm{\Tilde{x}}_M),
    \end{aligned}
\end{equation}
where $ \bmx_{\sigma_{i}} \in \mathbb{R}^{M\times d} $ denotes the output of the $ i $-th Gaussian block and $ M $ corresponds to the number of time points (i.e., clips or frames). $ w_i \in\mathbb{R}^{M} $ represents the aggregation weights for the $ i $-th Gaussian block, and $ \tau $ is the temperature factor. $ \bm{\Tilde{x}_j} \in\mathbb{R}^{d} $ denotes the aggregated feature at time point $ j $, while $ \bmx_{\text{MAIM}} $ is the final output.

\subsection{Learning Objectives}
Given the partial relevance in PRVR, where each video fully entails its corresponding text, a partial order relationship is established, with the text-query semantically subsumed by the video: $\text{text} \prec \text{video}$. Inspired by MERU \cite{meru}, we propose the Partial Order Preservation Loss to enforce this  relationship in Hyperbolic Space.
Given $ \bmV_f $ and $ \bmV_c $ from \cref{sec:overview}, a simple attention module similar to \cref{eq:simpleattention} is applied, followed by mean pooling to get the unified video representation $ \bmV_v $. The video and text representations are then mapped to Lorentz space via the exponential map, yielding $ \bmv, \bmt  \in \slorentz$,  as shown in \cref{fig:arc}(c). 
We define an entailment cone for each $ \bmv $, which is characterized by the half-aperture:
\begin{equation}
    \textbf{HA}({\bmv}) = \arcsin \left( \frac{2c}{  \lVert \bm{v_{s}} \rVert} \right).
\end{equation}
$ c = 0.1 $ is used to define the boundary conditions near the origin. 
We measure the exterior angle $ \textbf{EA}(\bmv, \bmt) = \pi - \angle O v t $ to penalize cases where  $ \bmt $ falls outside the entailment cone:
\begin{equation}
    \def\cvl{c \; \langle v,t \rangle _\mathcal{L}}
    \textbf{EA}(\bmv, \bmt) = \arccos \left( \frac{t_{0} + v_{0} \linner{v}{t}}{\lVert \bm{v_{s}} \rVert \sqrt{\left( \linner{v}{t}\right)^2 - 1}} \right).
\end{equation}
The Loss for a single video-text pair is given by:
\begin{equation}\label{eqn:meru_entailment}
    L_{pop}(\bmv, \bmt) = \max(0, \; \textbf{EA}(\bmv, \bmt) - \textbf{HA}(\bmv)).
\end{equation}

Besides, following MS-SL \cite{ms-sl}, we use the standard similarity retrieval loss to train the model, denoted as $L_{sim}$. Meanwhile, the query diversity \cite{GMMFORMER} $L_{div}$ is  used to  improve retrieval performance. The  aggregate loss is defined as:
\begin{equation}
L_{agg} =L_{sim} +  \lambda_1 L_{div} + \lambda_2 L_{pop},
\end{equation}
$\lambda_1$  and $\lambda_2$ are hyper-parameters that balance learning losses.

%% file: sections/Experiments.tex
\begin{table*}[t]
\centering
\resizebox{\textwidth}{!}{
\begin{tabular}{l ccccc c ccccc c  ccccc}
\toprule
\multirow{2}{*}{\textbf{Model}}& \multicolumn{5}{c}{\textbf{ActivityNet Captions}} &  & \multicolumn{5}{c}{\textbf{Charades-STA}}&& \multicolumn{5}{c}{\textbf{TVR}} \\
\cmidrule{2-6} \cmidrule{8-12} \cmidrule{14-18}
& \textbf{R@1} & \textbf{R@5} & \textbf{R@10} & \textbf{R@100} & \textbf{SumR} && \textbf{R@1} & \textbf{R@5} & \textbf{R@10} & \textbf{R@100} & \textbf{SumR} & & \textbf{R@1} & \textbf{R@5} & \textbf{R@10} & \textbf{R@100} & \textbf{SumR} \\
\midrule
\midrule
\rowcolor[HTML]{F2F2F2} \textbf{\emph{T2VR}} & & & &&& &  & & & & &  & & & & & \\

HGR \cite{chen2020fine} & 4.0 & 15.0 & 24.8 & 63.2 & 107.0  && 1.2 & 3.8 & 7.3 & 33.4& 45.7 && 1.7 & 4.9 & 8.3 & 35.2 & 50.1\\
RIVRL \cite{dong2022reading} & 5.2 & 18.0 & 28.2 & 66.4 & 117.8 &&1.6 & 5.6 & 9.4 & 37.7  & 54.3 && 9.4 & 23.4 & 32.2 & 70.6 & 135.6  \\
 DE++ \cite{dong2021dual}  & 5.3 & 18.4 & 29.2 & 68.0 & 121.0  && 1.7 & 5.6 & 9.6 & 37.1& 54.1 && 8.8 & 21.9 & 30.2 & 67.4 & 128.3\\
 CE \cite{liu2019use}  & 5.5 & 19.1 & 29.9 & 71.1 & 125.6  && 1.3 & 4.5 & 7.3 & 36.0& 49.1 && 3.7 & 12.8 & 20.1 & 64.5 & 101.1\\
CLIP4Clip \cite{luo2022clip4clip} & 5.9 & 19.3 & 30.4 & 71.6 & 127.3 &&1.8 & 6.5 & 10.9 & 44.2 & 63.4 &&  9.9 & 24.3 & 34.3 & 72.5 & 141.0\\
Cap4Video  \cite{wu2023cap4video} & 6.3 & 20.4 & 30.9 & 72.6 & 130.2 &&1.9 & 6.7 & 11.3 & 45.0 & 65.0 &&  10.3 & 26.4 & 36.8 & 74.0 & 147.5\\
\midrule
\rowcolor[HTML]{F2F2F2} \textbf{\emph{VCMR}} & & & &&& &  & & & & &  & & & & & \\
ReLoCLNet \cite{zhang2021video} & 5.7 & 18.9 & 30.0 & 72.0 & 126.6 &&1.2 & 5.4 & 10.0 & 45.6 & 62.3 & & 10.0 & 26.5 & 37.3 & 81.3  & 155.1 \\
XML  \cite{lei2020tvr} &  5.3 & 19.4 & 30.6 & 73.1 & 128.4 &&1.6 & 6.0 & 10.1 & 46.9 & 64.6 && 10.7 & 28.1 & 38.1 & 80.3  & 157.1 \\ 
CONQUER \cite{hou2021conquer}&  6.5 & 20.4 & 31.8 & 74.3  & 133.1 &&1.8 & 6.3 & 10.3 & 47.5  & 66.0 &&11.0 & 28.9 & 39.6 & 81.3  & 160.8 \\ 
JSG \cite{jsg}&  6.8 & 22.7 & 34.8 & 76.1  & 140.5 &&2.4 & 7.7 & 12.8 & 49.8  & 72.7 &&- & - & - & -  & -\\ 
\midrule
\rowcolor[HTML]{F2F2F2} \textbf{\emph{PRVR}} & & & &&& &  & & & & &  & & & & & \\
MS-SL \cite{ms-sl}& 7.1 & 22.5 & 34.7 & 75.8 & 140.1 &&1.8 & 7.1 & 11.8 & 47.7 & 68.4& & 13.5 & 32.1 & 43.4 & 83.4 & 172.4       \\
 PEAN \cite{PEAN}&7.4 & 23.0 & 35.5 & 75.9 & 141.8 &&  \textbf{2.7} & 8.1 & 13.5 & 50.3 & 74.7 && 13.5 & 32.8 & 44.1 & 83.9 & 174.2 \\
  LH \cite{lh}& 7.4 & 23.5 & 35.8 & 75.8 & 142.4 &&2.1 & 7.5 & 12.9 & 50.1 & 72.7& & 13.2 & 33.2 & 44.4 & 85.5 & 176.3 \\
 BGM-Net \cite{bgmnet}& 7.2 & 23.8 & 36.0 & 76.9 & 143.9 &&1.9 & 7.4 & 12.2 & 50.1 & 71.6 && 14.1 & 34.7 & 45.9 & 85.2 & 179.9 \\
 GMMFormer \cite{GMMFORMER}&8.3 & 24.9 & 36.7 & 76.1 & 146.0 && 2.1 & 7.8 & 12.5 & 50.6 & 72.9 &&13.9 & 33.3 & 44.5 & 84.9 & 176.6 \\
 DL-DKD \cite{DKD}& 8.0 & 25.0 & 37.5 & 77.1 & 147.6 &&- & - & - & - & - && 14.4 & 34.9 & 45.8 & 84.9 & 179.9  \\
\rowcolor[HTML]{ADD8E6} 
 \textbf{\modelname{} (ours)} & \textbf{8.7} & \textbf{27.1} & \textbf{40.1} & \textbf{79.0} & \textbf{154.9} &&\textbf{2.6} & \textbf{8.5} & \textbf{13.7} & \textbf{54.0} & \textbf{78.7} &&   \textbf{15.7} & \textbf{37.1} & \textbf{48.5} & \textbf{86.4} & \textbf{187.7}     \\
\hline
\bottomrule
\end{tabular}
}
\caption{\textbf{Retrieval performance of \modelname{} and other faithfull methods on ActivityNet Captions, Charades-STA and TVR.} 
State-of-the-art performance is highlighted in \textbf{bold}. “-” indicates that the corresponding results are unavailable.}
\label{tab:main}
\end{table*}
\section{Experiments} \label{sec:experiments}
\subsection{Experimental Setup} \label{subsec:exp_setup}
\noindent \textbf{Datasets} \quad We conduct experiments on three benchmark  datasets: (\textbf{i}) \textbf{ActivityNet Captions} \cite{krishna2017dense}, which comprises approximately 20K YouTube videos with an average duration of 118 seconds. Each video contains an average of 3.7 annotated moments with corresponding textual descriptions. (\textbf{ii}) \textbf{TV show Retrieval (TVR)} \cite{lei2020tvr}, consisting of 21.8K videos sourced from six TV shows. Each video is associated with five natural language descriptions covering different moments.  (\textbf{iii}) \textbf{Charades-STA} \cite{gao2017tall}, which includes 6,670 videos annotated with 16,128 sentence descriptions. On average, each video contains approximately 2.4 moments with corresponding textual queries. We adopt the same data split as used in prior studies\cite{ms-sl,GMMFORMER}. It is important to note that the moment annotations  are unavailable in the PRVR task.

\noindent \textbf{Metrics} \quad Following previous works \cite{ms-sl,GMMFORMER},  we adopt rank-based evaluation metrics, specifically $R$@$K$ ($K$ = 1, 5, 10, 100). The metric $R$@$K$ represents the proportion of queries for which the correct item appears within the top $K$ positions of the ranking list. All results are reported as percentages ($\%$), where higher values indicate superior retrieval performance. To facilitate an overall comparison, we also report the Sum of all Recalls (SumR).

\subsection{Implementation Details}
\noindent \textbf{Data Processing} \quad For video representations on TVR, we employ the feature set provided by \citet{lei2020tvr}, which comprises 3,072-dimensional visual features obtained by concatenating frame-level ResNet152 features \cite{he2016deep} and segment-level I3D features \cite{carreira2017quo}. For ActivityNet Captions and Charades-STA, we only utilize I3D features as provided by \citet{zhang2020hierarchical} and \citet{mun2020local}, respectively.   For sentence representations, we adopt the 768-dimensional RoBERTa features supplied by \citet{lei2020tvr} for TVR. On ActivityNet Captions and Charades-STA, we employ 1,024-dimensional RoBERTa features extracted using MS-SL\cite{ms-sl}.

\noindent \textbf{Model Configurations} \quad 
The \modelname{} block consists of 8 Gaussian blocks ($N_O = 8$), 4 Lorentz Attention blocks ($N_\mathcal{L} = 4$), with Gaussian variances ranging from $2^1$ to $2^{N_\mathcal{L}-1}$ and $\infty$, and 4 Euclidean Attention blocks ($N_E = 4$), with Gaussian variances ranging from $2^1$ to $2^{N_E-1}$ and $\infty$. The latent dimension  $ d = 384 $ with 4 attention heads.

\noindent \textbf{Training Configurations} \quad We employ the Adam optimizer with a mini-batch size of 128 and set the number of epochs to 100. The model is implemented using PyTorch and trained on one Nvidia RTX 3080 Ti GPU.
We adopt a learning rate adjustment schedule similar to  MS-SL. 

\begin{table*}[t]
\centering
\resizebox{\textwidth}{!}{
\begin{tabular}{cl ccccc c ccccc c  ccccc}
\toprule
\multirow{2}{*}{ID}&\multirow{2}{*}{\textbf{Model}}& \multicolumn{5}{c}{\textbf{ActivityNet Captions}} &  & \multicolumn{5}{c}{\textbf{Charades-STA}}&& \multicolumn{5}{c}{\textbf{TVR}} \\
\cmidrule{3-7} \cmidrule{9-13} \cmidrule{15-19}
&& \textbf{R@1} & \textbf{R@5} & \textbf{R@10} & \textbf{R@100} & \textbf{SumR} && \textbf{R@1} & \textbf{R@5} & \textbf{R@10} & \textbf{R@100} & \textbf{SumR} & & \textbf{R@1} & \textbf{R@5} & \textbf{R@10} & \textbf{R@100} & \textbf{SumR} \\
\midrule
\midrule
\rowcolor[HTML]{ADD8E6} 
 (0) & \textbf{\modelname{} (full)} & \textbf{8.7} & \textbf{27.1} & \textbf{40.1} & \textbf{79.0} & \textbf{154.9} &&\textbf{2.6} & \textbf{8.5} & \textbf{13.7} & \textbf{54.0} & \textbf{78.7} &&   \textbf{15.7} & \textbf{37.1} & \textbf{48.5} & \textbf{86.4} & \textbf{187.7}     \\
\midrule
\rowcolor[HTML]{F2F2F2} \multicolumn{19}{l}{\textbf{\emph{Efficacy of Multi-scale Branches}}} \\
(1)& $w/o$ gaze branch& 7.6& 24.4 & 36.7 & 77.3 & 146.1&&1.8 & 8.0 & 13.9 & 50.8  & 74.5 &&  13.9& 34.0 & 45.2&85.3 & 178.3 \\
(2)& $w/o$ glance branch  & 6.4 & 21.7 & 33.6 & 75.4 & 137.2  &&1.6 & 7.7 & 13.1 & 48.4 & 70.8 &&  11.4 & 30.5 & 41.8 & 82.4 & 166.1 \\
\midrule
\rowcolor[HTML]{F2F2F2} \multicolumn{19}{l}{\textbf{\emph{Efficacy of Different Loss Terms}}} \\
(3) & $L_{sim}$ Only & 7.7 & 25.0 & 38.1 & 78.3 & 149.1 &&2.0 & 8.1 & 13.2 & 52.0 & 75.3 & & 15.1 & 36.2& 47.8 & 86.0  & 185.2\\
(4) & $w/o$ $L_{div}$  &  8.5 & 26.6 & 39.6 & 78.8 & 153.5 &&2.0 & 7.8 & 13.6 & 53.0 & 76.4 && 15.7 & 36.4 & 48.4 & 86.0  & 186.5\\ 
(5) &$w/o$ $ L_{pop}$&  8.6 & 26.9 & 39.7 & 78.8 & 154.0  &&2.2 & 8.4 & 14.0 & 53.0  & 77.6 &&15.6 & 36.8 & 48.4 & 86.0  & 186.8\\ 
\midrule
\rowcolor[HTML]{F2F2F2} \multicolumn{19}{l}{\textbf{\emph{Efficacy of various Aggregation Strategies}}} \\
(6)& $w/$ $ \mathbf{MP}$& 8.5 & 25.7 & 38.2 & 77.8 & 150.2 &&2.0 & 8.0 & 13.2 & 52.1 & 75.3& & 15.2 & 36.5 & 47.4 & 86.0 & 185.1      \\
(7)& $w/$ $ \mathbf{CL}$&8.7 & 26.8 & 39.5 & 78.6 & 153.6 &&  2.0 & 8.2 & 13.9 & 52.0 & 76.1 && 15.3 & 36.9 & 48.4 & 86.0 & 186.6 \\
\hline
\bottomrule
\end{tabular}
}
 \caption{Ablation Study of \modelname{}.  The best scores are marked in \textbf{bold}.}
 \label{tab:ablation}
\end{table*}
\subsection{Comparison with State-of-the arts}
\noindent \textbf{Baselines} \quad We select six representative PRVR baselines for comparison: MS-SL \cite{ms-sl}, PEAN \cite{PEAN}, LH \cite{lh}, BGM-Net \cite{bgmnet}, GMMFormer \cite{GMMFORMER}, and DL-DKD \cite{DKD}. 
% We have discussed them in the Related Works section. 
 We also compare \modelname{} with methods for T2VR and VCMR. For T2VR, we select
 six T2VR models: CE \cite{liu2019use}, 
 % HTM \cite{miech2019howto100m}, 
 HGR \cite{chen2020fine}, DE++ \cite{dong2021dual}, RIVRL \cite{dong2022reading}, CLIP4Clip \cite{luo2022clip4clip}, Cap4Video \cite{wu2023cap4video}, 
For VCMR, we consider four models: XML \cite{lei2020tvr}, ReLoCLNet \cite{zhang2021video}, CONQUER \cite{hou2021conquer} and JSG\cite{jsg}.

\noindent \textbf{Retrieval Performance} \quad
\cref{tab:main} presents the retrieval performance of various models on three large-scale video datasets. As observed, T2VR models, designed to capture overall video-text relevance, underperform  for PRVR. VCMR models, which focus on moment retrieval, achieve better results. PRVR methods perform best as they are specifically designed for this task. Attributed to hyperbolic space learning and effective utilization of video hierarchical structure priors, \modelname{} consistently surpasses all baselines. It outperforms DL-DKD by $\mathbf{4.9\%}$ and $\mathbf{4.3\%}$ in \textbf{SumR} on ActivityNet Captions and TVR, respectively, and exceeds PEAN by $\mathbf{5.4\%}$ on Charades-STA.

\subsection{Model Analyses}
\begin{figure}[t]
\centering
    {
        \includegraphics[width = 0.2200\textwidth]{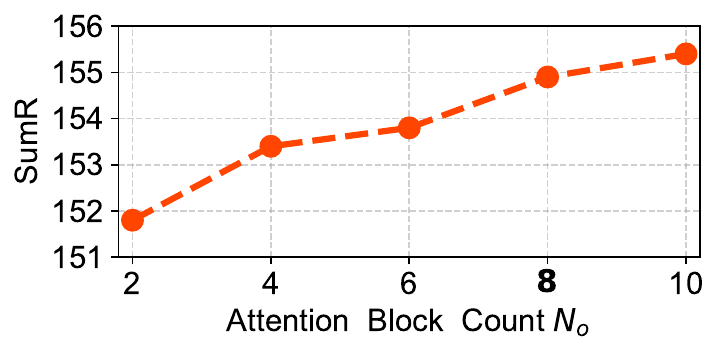}
    }
    {
        \includegraphics[width = 0.2200\textwidth]{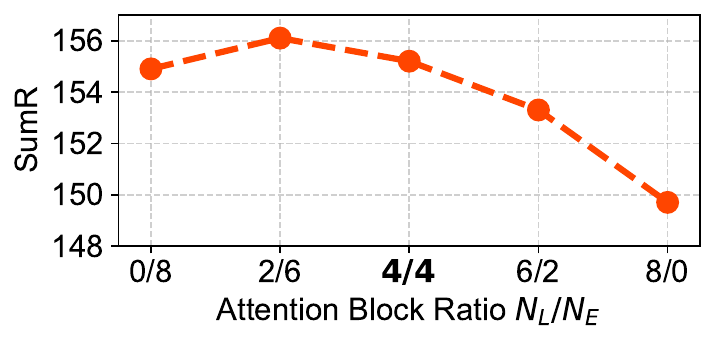}
    }\\
    \scalebox{0.8}{\small {(a) ActivityNet Captions}} \\
    {
        \includegraphics[width = 0.2200\textwidth]{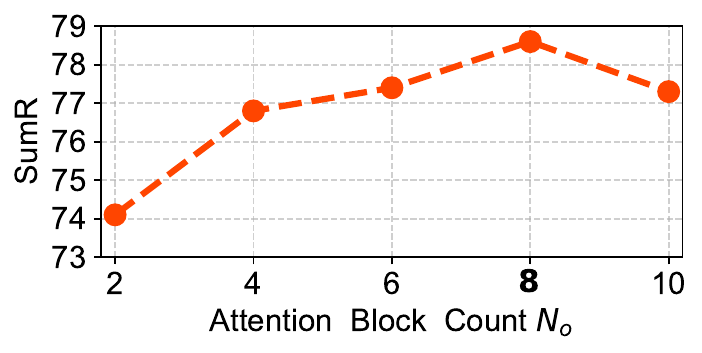}
    }
    {
        \includegraphics[width = 0.2200\textwidth]{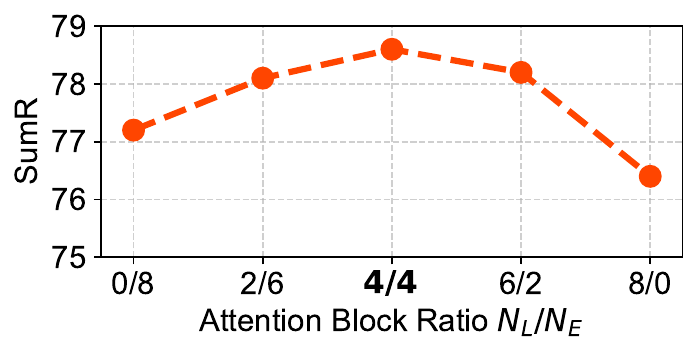}
    }\\
    \scalebox{0.8}{{(b) Charades-STA}} \\
    {
        \includegraphics[width = 0.2200\textwidth]{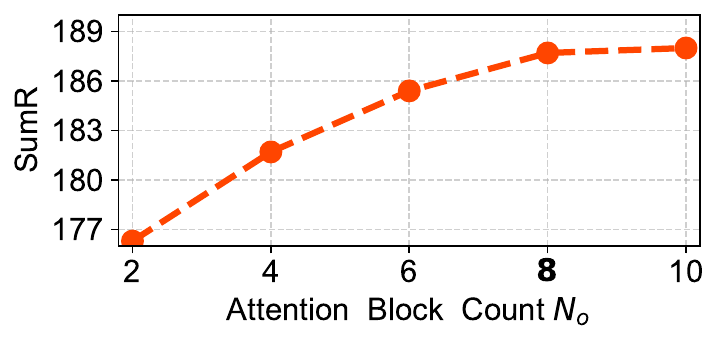}
    }
    {
        \includegraphics[width = 0.2200\textwidth]{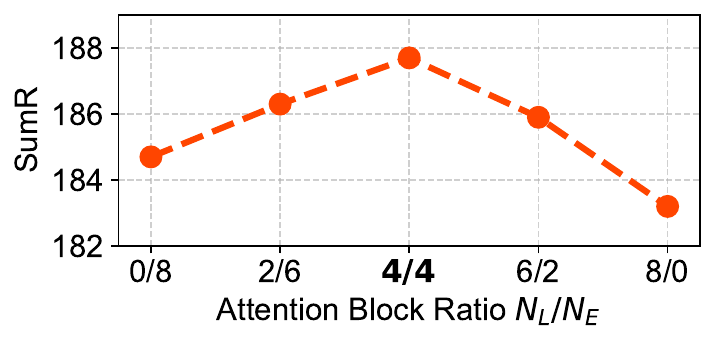}
    }\\
    \scalebox{0.9}{\small {(c) TVR}} \\
    \caption{The influence of different attention blocks, with default settings marked in bold.}
    \label{fig:attention}
\end{figure}

\noindent \textbf{Efficacy of Temporal Modeling Design} \quad
We perform ablation studies to examine the effect of the attention block number $N_o$ and the attention mechanism ratio $N_L/N_E$, with results shown in \cref{fig:attention}. Model performance improves as $N_o$ increases, then stabilizes or declines when $N_o \geq 8$. Even with only two attention blocks, \modelname{} surpasses most competing methods.
Furthermore, using solely Euclidean or Lorentz attention blocks results in suboptimal performance, whereas the hybrid attention block achieves the best results. This may be attributed to the differences in representational focus: Euclidean space emphasizes fine-grained local feature learning and sometimes overlooks global hierarchical structures, while hyperbolic space prioritizes global hierarchical relationships at the expense of local details. Moreover, hyperbolic space tends to be more sensitive to noise and numerically unstable.
By integrating hybrid spaces, \modelname{} achieves mutual compensation, enhancing representation learning and facilitating video  semantic understanding.

\begin{figure}[t]
\centering
    \setcounter{subfigure}{0}
  \subfloat[\small{$w/o$ Lorentz Attention}]{\includegraphics[width = 0.2200\textwidth]{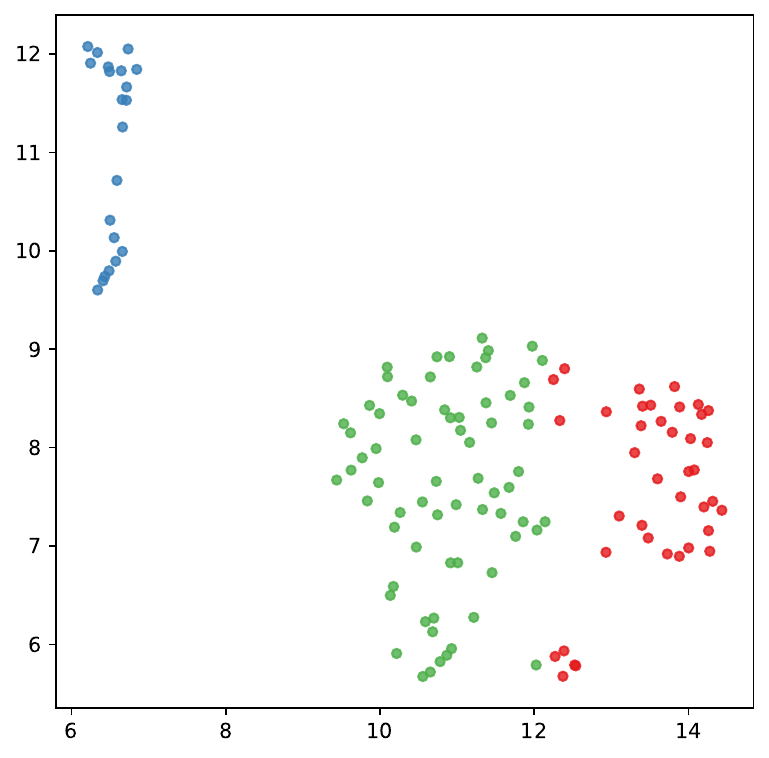}}
  \subfloat[\small{$w/$ Lorentz Attention}]{\includegraphics[width = 0.2200\textwidth]{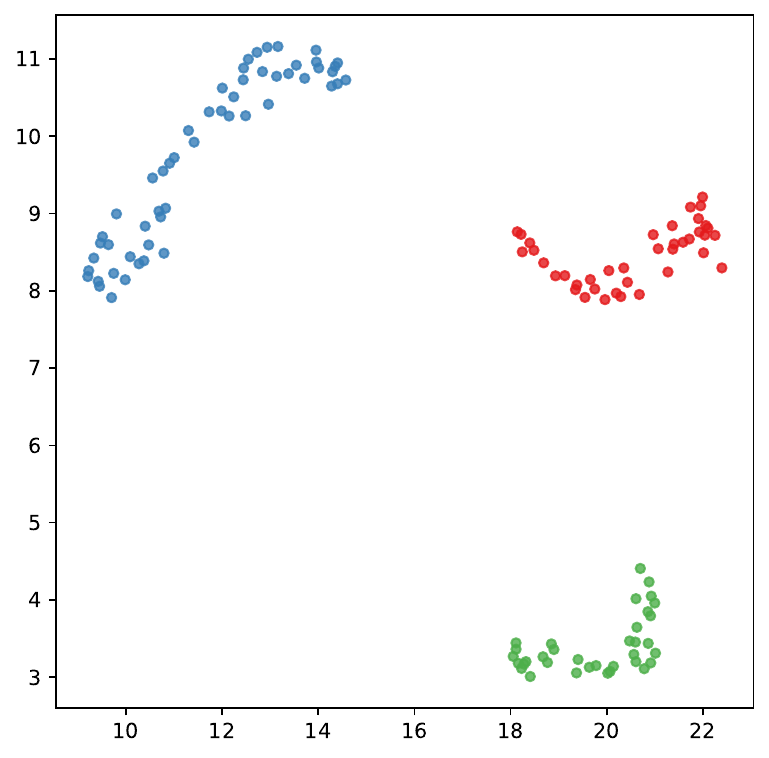}}
  \caption{The UMAP \cite{umap} visualization displays the learned frame embeddings  from a video in TVR. Data points of the same color correspond to the same moment. }
  \label{fig:umap}
\end{figure}
\noindent \textbf{Efficacy of Hyperbolic Learning} \quad
Hyperbolic learning demonstrates significant advantages in capturing the hierarchical structure of videos. As illustrated in \cref{fig:umap}(a), embeddings learned solely in Euclidean space exhibit indistinct cluster boundaries, with red and green points at the periphery closely interspersed.
In contrast, \cref{fig:umap}(b) demonstrates that incorporating Lorentz attention facilitates the learning of more discriminative representations, while  refining moment cluster boundaries, increasing inter-moment separation, and compacting intra-moment frame distributions, revealing a more pronounced hierarchical structure.

\noindent \textbf{Efficacy of Multi-scale Branches} \quad
To evaluate the effectiveness of the multi-scale branches, we conduct comparative experiments by removing either the glance clip-level branch or the gaze frame-level branch. As shown in \cref{tab:ablation}, the absence of any branch leads to a noticeable performance degradation. These results not only validate the efficacy of the coarse-to-fine multi-granularity retrieval mechanism but also highlight the complementary nature of the two branches.  
\begin{figure}[t]
\centering
    {
        \includegraphics[width = 0.4800\textwidth]{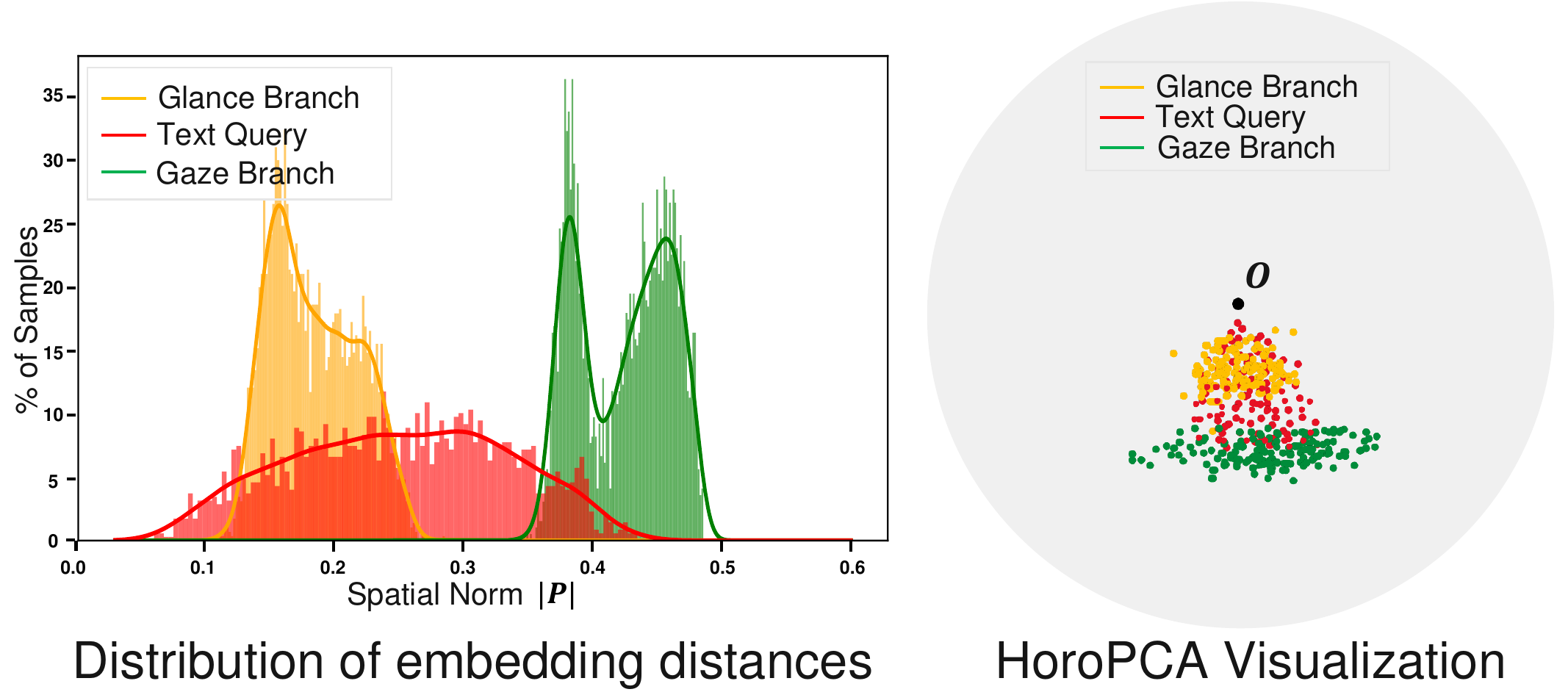}
    }\\
    \scalebox{0.9}{\small {(a) $w/o$ $L_{pop}$}} \\
    % Second row with reduced font size for the title
    {
        \includegraphics[width = 0.4800\textwidth]{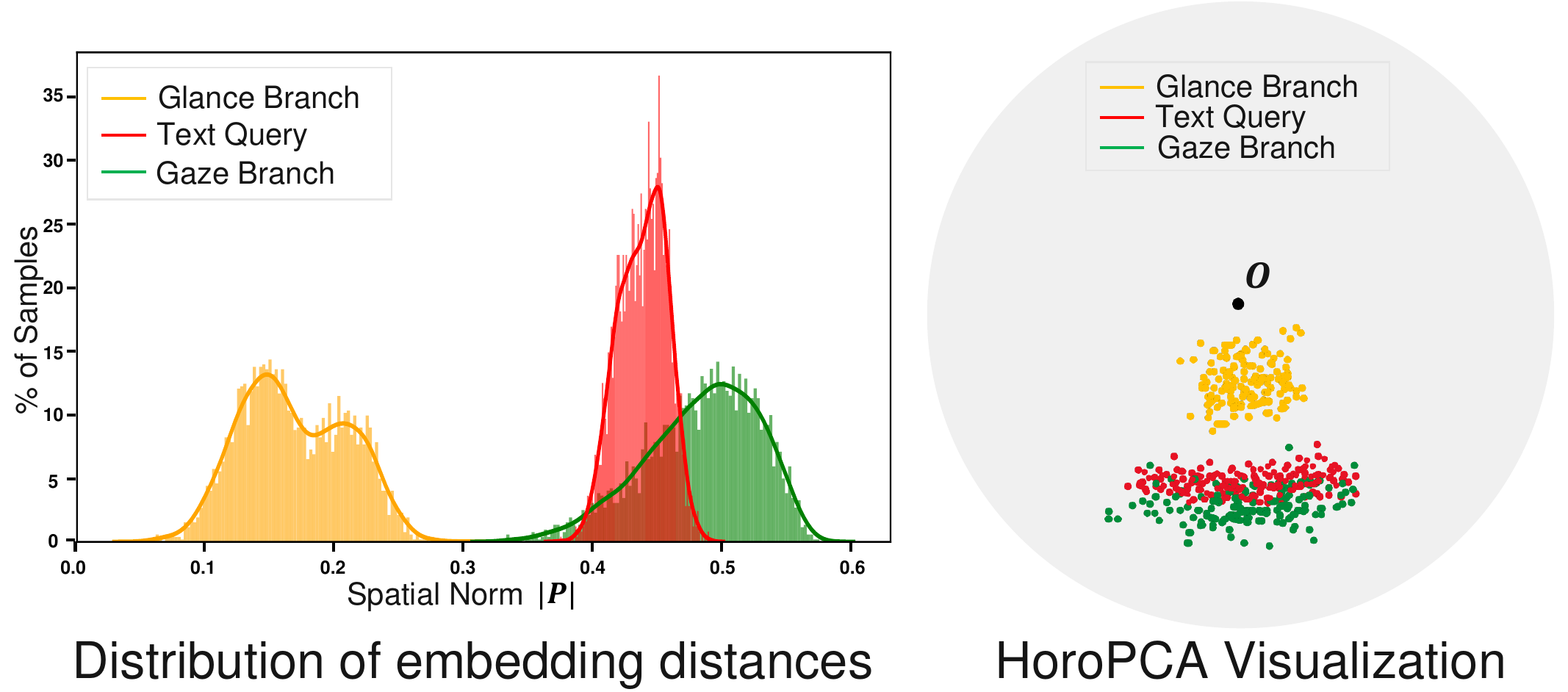}
    }\\
   \scalebox{0.9}{\small {(b) $w/$ $L_{pop}$}} \\
    \caption{Visualization of the learned hyperbolic space. The closer to the origin, the higher semantic hierarchy and coarser granularity.}
    \label{fig:vis2}
\end{figure}
\noindent \textbf{Efficacies of Different Loss Terms} \quad To analyze the effectiveness of three loss terms (\ie$L_{sim}$, $L_{div}$ and $L_{pop}$) of \modelname{}, we construct several \modelname{} variants: 
(\textbf{i}) $L_{sim}$ Only: train the model with merely $L_{sim}$. 
(\textbf{ii}) w/o $L_{div}$: We train the model without query diverse  learning. 
(\textbf{iii}) w/o $L_{pop}$: \modelname{} removes  the partial order preservation task.
As shown in \cref{tab:ablation}, the worst performance occurs when only $L_{sim}$ is used. Comparing Variant (5) with Variant (3), adding  $L_{div}$  increases the SumR, which can validate its necessity. Similarly, comparing Variant (4) with Variant (3)  and  \cref{fig:vis2},  integrating $L_{pop}$ not only boosts retrieval accuracy but also ensures that the text query remains semantically embedded within the corresponding video, preserving partial relevance.

\noindent \textbf{Efficacy of Aggregation Strategy} \quad 
 We compare three aggregation strategies:  
(\textbf{i}) $w/$ $\mathbf{MP}$: mean pooling for static fusion.  
(\textbf{ii}) $w/$ $\mathbf{CL}$: feature concatenation with linear layers .  
(\textbf{iii}) MAIM (default):  mean-guided adaptive interaction module. 
As shown in \cref{tab:ablation}, MP performs the worst due to its fixed static fusion, which limits semantic interaction. CL improves upon MP by leveraging linear layers for dynamic feature fusion. MAIM achieves the best performance by learning adaptive aggregation weights and dynamically selecting hyperbolic information under global guidance.  

\noindent \textbf{Visualization of Hyperbolic Space} \quad 
Inspired by HyCoCLIP \cite{hycoclip}, we visualize the learned hyperbolic space by sampling 3K embeddings from the TVR training set. We analyze their norm distribution via histogram and reduce dimensionality using HoroPCA \cite{horopca}, as shown in \cref{fig:vis2}.
 Glance branch embeddings are positioned closer to the origin than text query embeddings, indicating that clip-level video representations subsume textual queries. This phenomenon can be attributed to $L_{pop}$, which enforces the partial order relationship between video and text representations. In contrast, without $L_{pop}$, embeddings exhibit uncorrelated distributions. Moreover, text queries, being coarser in semantics, lie closer to the origin than fine-grained gaze-level embeddings, reflecting a clear hierarchical structure.

%% file: sections/Conclusions.tex
\section{Conclusions}
\label{sec:conclusion}
In this paper, we propose \modelname{}, a novel hyperbolic modeling framework tailored for PRVR. By leveraging the intrinsic geometric properties of hyperbolic space, \modelname{} effectively captures the hierarchical and multi-granular structure of untrimmed videos, thereby enhancing video-text retrieval accuracy.  Furthermore, to ensure partial  relevance between paired videos and text,  a partial order preservation loss is introduced to  enforce their semantic entailment.
Extensive experiments indicate that \modelname{} consistently outperforms state-of-the-art methods. Our study offers a new perspective for PRVR with hyperbolic learning, which we hope will inspire further research in this direction.  

\paragraph{Acknowledgments}
We sincerely thank the anonymous reviewers and chairs for their efforts and constructive suggestions, which have greatly helped us improve the manuscript. 
This work is supported in part by the National Natural Science Foundation of China under grant 624B2088, 62171248, 62301189, the PCNL KEY project (PCL2023AS6-1), and Shenzhen Science and Technology Program under Grant KJZD20240903103702004, JCYJ20220818101012025, GXWD20220811172936001.
Long Chen was supported by the Hong Kong SAR RGC Early Career Scheme (26208924), the National Natural Science Foundation of China Young Scholar Fund (62402408), Huawei Gift Fund, and the HKUST Sports Science and Technology Research Grant (SSTRG24EG04).

%% file: main.bbl
\begin{thebibliography}{68}
\providecommand{\natexlab}[1]{#1}
\providecommand{\url}[1]{\texttt{#1}}
\expandafter\ifx\csname urlstyle\endcsname\relax
  \providecommand{\doi}[1]{doi: #1}\else
  \providecommand{\doi}{doi: \begingroup \urlstyle{rm}\Url}\fi

\bibitem[Atigh et~al.(2022)Atigh, Schoep, Acar, Van~Noord, and Mettes]{segmenation1}
Mina~Ghadimi Atigh, Julian Schoep, Erman Acar, Nanne Van~Noord, and Pascal Mettes.
\newblock Hyperbolic image segmentation.
\newblock In \emph{Proceedings of the IEEE/CVF conference on computer vision and pattern recognition}, pages 4453--4462, 2022.

\bibitem[Carreira and Zisserman(2017)]{carreira2017quo}
Joao Carreira and Andrew Zisserman.
\newblock Quo vadis, action recognition? a new model and the kinetics dataset.
\newblock In \emph{proceedings of the IEEE Conference on Computer Vision and Pattern Recognition}, pages 6299--6308, 2017.

\bibitem[Chami et~al.(2021)Chami, Gu, Nguyen, and Ré]{horopca}
Ines Chami, Albert Gu, Dat Nguyen, and Christopher Ré.
\newblock Horopca: Hyperbolic dimensionality reduction via horospherical projections, 2021.

\bibitem[Chen et~al.(2023{\natexlab{a}})Chen, Peng, Cao, and R{\"o}ning]{segmentation2}
Bike Chen, Wei Peng, Xiaofeng Cao, and Juha R{\"o}ning.
\newblock Hyperbolic uncertainty aware semantic segmentation.
\newblock \emph{IEEE Transactions on Intelligent Transportation Systems}, 25\penalty0 (2):\penalty0 1275--1290, 2023{\natexlab{a}}.

\bibitem[Chen et~al.(2020)Chen, Zhao, Jin, and Wu]{chen2020fine}
Shizhe Chen, Yida Zhao, Qin Jin, and Qi Wu.
\newblock Fine-grained video-text retrieval with hierarchical graph reasoning.
\newblock In \emph{Proceedings of the IEEE/CVF Conference on Computer Vision and Pattern Recognition}, pages 10638--10647, 2020.

\bibitem[Chen et~al.(2021)Chen, Han, Lin, Zhao, Liu, Li, Sun, and Zhou]{net2}
Weize Chen, Xu Han, Yankai Lin, Hexu Zhao, Zhiyuan Liu, Peng Li, Maosong Sun, and Jie Zhou.
\newblock Fully hyperbolic neural networks.
\newblock \emph{arXiv preprint arXiv:2105.14686}, 2021.

\bibitem[Chen et~al.(2023{\natexlab{b}})Chen, Jiang, Xu, Cao, Mo, and Shen]{jsg}
Zhiguo Chen, Xun Jiang, Xing Xu, Zuo Cao, Yijun Mo, and Heng~Tao Shen.
\newblock Joint searching and grounding: Multi-granularity video content retrieval.
\newblock In \emph{Proceedings of the 31st ACM International Conference on Multimedia}, pages 975--983, 2023{\natexlab{b}}.

\bibitem[Cheng et~al.(2024)Cheng, Kong, Jiang, and Guo]{t-d3n}
Dingxin Cheng, Shuhan Kong, Bin Jiang, and Qiang Guo.
\newblock Transferable dual multi-granularity semantic excavating for partially relevant video retrieval.
\newblock \emph{Image and Vision Computing}, 149:\penalty0 105168, 2024.

\bibitem[Cho et~al.(2025)Cho, Moon, Jun, Jung, and Heo]{cho2025ambiguity}
Cheol-Ho Cho, WonJun Moon, Woojin Jun, MinSeok Jung, and Jae-Pil Heo.
\newblock Ambiguity-restrained text-video representation learning for partially relevant video retrieval.
\newblock In \emph{Proceedings of the AAAI Conference on Artificial Intelligence}, pages 2500--2508, 2025.

\bibitem[Desai et~al.(2023)Desai, Nickel, Rajpurohit, Johnson, and Vedantam]{meru}
Karan Desai, Maximilian Nickel, Tanmay Rajpurohit, Justin Johnson, and Ramakrishna Vedantam.
\newblock {Hyperbolic Image-Text Representations}.
\newblock In \emph{Proceedings of the International Conference on Machine Learning}, 2023.

\bibitem[Dong et~al.(2018)Dong, Li, and Snoek]{dong2018predicting}
Jianfeng Dong, Xirong Li, and Cees~GM Snoek.
\newblock Predicting visual features from text for image and video caption retrieval.
\newblock \emph{IEEE Transactions on Multimedia}, 20\penalty0 (12):\penalty0 3377--3388, 2018.

\bibitem[Dong et~al.(2019)Dong, Li, Xu, Ji, He, Yang, and Wang]{dong2019dual}
Jianfeng Dong, Xirong Li, Chaoxi Xu, Shouling Ji, Yuan He, Gang Yang, and Xun Wang.
\newblock Dual encoding for zero-example video retrieval.
\newblock In \emph{Proceedings of the IEEE/CVF conference on computer vision and pattern recognition}, pages 9346--9355, 2019.

\bibitem[Dong et~al.(2021)Dong, Li, Xu, Yang, Yang, Wang, and Wang]{dong2021dual}
Jianfeng Dong, Xirong Li, Chaoxi Xu, Xun Yang, Gang Yang, Xun Wang, and Meng Wang.
\newblock Dual encoding for video retrieval by text.
\newblock \emph{IEEE Transactions on Pattern Analysis and Machine Intelligence}, 44\penalty0 (8):\penalty0 4065--4080, 2021.

\bibitem[Dong et~al.(2022{\natexlab{a}})Dong, Chen, Zhang, Yang, Chen, Li, and Wang]{ms-sl}
Jianfeng Dong, Xianke Chen, Minsong Zhang, Xun Yang, Shujie Chen, Xirong Li, and Xun Wang.
\newblock Partially relevant video retrieval.
\newblock In \emph{Proceedings of the 30th ACM International Conference on Multimedia}, pages 246--257, 2022{\natexlab{a}}.

\bibitem[Dong et~al.(2022{\natexlab{b}})Dong, Wang, Chen, Qu, Li, He, and Wang]{dong2022reading}
Jianfeng Dong, Yabing Wang, Xianke Chen, Xiaoye Qu, Xirong Li, Yuan He, and Xun Wang.
\newblock Reading-strategy inspired visual representation learning for text-to-video retrieval.
\newblock \emph{IEEE Transactions on Circuits and Systems for Video Technology}, 32\penalty0 (8):\penalty0 5680--5694, 2022{\natexlab{b}}.

\bibitem[Dong et~al.(2023)Dong, Zhang, Zhang, Chen, Liu, Qu, Wang, and Liu]{DKD}
Jianfeng Dong, Minsong Zhang, Zheng Zhang, Xianke Chen, Daizong Liu, Xiaoye Qu, Xun Wang, and Baolong Liu.
\newblock Dual learning with dynamic knowledge distillation for partially relevant video retrieval.
\newblock In \emph{Proceedings of the IEEE/CVF International Conference on Computer Vision}, pages 11302--11312, 2023.

\bibitem[Ermolov et~al.(2022)Ermolov, Mirvakhabova, Khrulkov, Sebe, and Oseledets]{hyperbolicvisiontransformer}
Aleksandr Ermolov, Leyla Mirvakhabova, Valentin Khrulkov, Nicu Sebe, and Ivan Oseledets.
\newblock Hyperbolic vision transformers: Combining improvements in metric learning.
\newblock In \emph{2022 IEEE/CVF Conference on Computer Vision and Pattern Recognition (CVPR)}, pages 7399--7409, 2022.

\bibitem[Faghri et~al.(2017)Faghri, Fleet, Kiros, and Fidler]{faghri2017vse++}
Fartash Faghri, David~J Fleet, Jamie~Ryan Kiros, and Sanja Fidler.
\newblock Vse++: Improving visual-semantic embeddings with hard negatives.
\newblock \emph{arXiv preprint arXiv:1707.05612}, 2017.

\bibitem[Fang et~al.(2025)Fang, Zhou, Kong, Gao, Chen, Liang, Ma, and Xia]{fang2025grounding}
Hao Fang, Changle Zhou, Jiawei Kong, Kuofeng Gao, Bin Chen, Tao Liang, Guojun Ma, and Shu-Tao Xia.
\newblock Grounding language with vision: A conditional mutual information calibrated decoding strategy for reducing hallucinations in lvlms.
\newblock \emph{arXiv preprint arXiv:2505.19678}, 2025.

\bibitem[Fang et~al.(2024)Fang, Dang, Wang, and Huang]{lh}
Sheng Fang, Tiantian Dang, Shuhui Wang, and Qingming Huang.
\newblock Linguistic hallucination for text-based video retrieval.
\newblock \emph{IEEE Transactions on Circuits and Systems for Video Technology}, 34\penalty0 (10):\penalty0 9692--9705, 2024.

\bibitem[Ganea et~al.(2018{\natexlab{a}})Ganea, Becigneul, and Hofmann]{entail}
Octavian Ganea, Gary Becigneul, and Thomas Hofmann.
\newblock Hyperbolic entailment cones for learning hierarchical embeddings.
\newblock In \emph{Proceedings of the 35th International Conference on Machine Learning}, pages 1646--1655. PMLR, 2018{\natexlab{a}}.

\bibitem[Ganea et~al.(2018{\natexlab{b}})Ganea, B{\'e}cigneul, and Hofmann]{net1}
Octavian Ganea, Gary B{\'e}cigneul, and Thomas Hofmann.
\newblock Hyperbolic neural networks.
\newblock \emph{Advances in neural information processing systems}, 31, 2018{\natexlab{b}}.

\bibitem[Gao et~al.(2017)Gao, Sun, Yang, and Nevatia]{gao2017tall}
Jiyang Gao, Chen Sun, Zhenheng Yang, and Ram Nevatia.
\newblock Tall: Temporal activity localization via language query.
\newblock In \emph{Proceedings of the IEEE international conference on computer vision}, pages 5267--5275, 2017.

\bibitem[He et~al.(2016)He, Zhang, Ren, and Sun]{he2016deep}
Kaiming He, Xiangyu Zhang, Shaoqing Ren, and Jian Sun.
\newblock Deep residual learning for image recognition.
\newblock In \emph{Proceedings of the IEEE conference on computer vision and pattern recognition}, pages 770--778, 2016.

\bibitem[He et~al.(2024)He, Yang, and Ying]{net4}
Neil He, Menglin Yang, and Rex Ying.
\newblock Lorentzian residual neural networks.
\newblock \emph{arXiv preprint arXiv:2412.14695}, 2024.

\bibitem[Hou et~al.(2021)Hou, Ngo, and Chan]{hou2021conquer}
Zhijian Hou, Chong-Wah Ngo, and Wing~Kwong Chan.
\newblock Conquer: Contextual query-aware ranking for video corpus moment retrieval.
\newblock In \emph{Proceedings of the 29th ACM International Conference on Multimedia}, pages 3900--3908, 2021.

\bibitem[Jiang et~al.(2023)Jiang, Chen, Xu, Shen, Cao, and Cai]{PEAN}
Xun Jiang, Zhiguo Chen, Xing Xu, Fumin Shen, Zuo Cao, and Xunliang Cai.
\newblock Progressive event alignment network for partial relevant video retrieval.
\newblock In \emph{2023 IEEE International Conference on Multimedia and Expo (ICME)}, pages 1973--1978. IEEE, 2023.

\bibitem[Khrulkov et~al.(2020)Khrulkov, Mirvakhabova, Ustinova, Oseledets, and Lempitsky]{Khrulkov_2020_CVPR}
Valentin Khrulkov, Leyla Mirvakhabova, Evgeniya Ustinova, Ivan Oseledets, and Victor Lempitsky.
\newblock Hyperbolic image embeddings.
\newblock In \emph{Proceedings of the IEEE/CVF Conference on Computer Vision and Pattern Recognition (CVPR)}, 2020.

\bibitem[Krishna et~al.(2017)Krishna, Hata, Ren, Fei-Fei, and Carlos~Niebles]{krishna2017dense}
Ranjay Krishna, Kenji Hata, Frederic Ren, Li Fei-Fei, and Juan Carlos~Niebles.
\newblock Dense-captioning events in videos.
\newblock In \emph{Proceedings of the IEEE international conference on computer vision}, pages 706--715, 2017.

\bibitem[Law et~al.(2019)Law, Liao, Snell, and Zemel]{lorentizan}
Marc Law, Renjie Liao, Jake Snell, and Richard Zemel.
\newblock Lorentzian distance learning for hyperbolic representations.
\newblock In \emph{Proceedings of the 36th International Conference on Machine Learning}, pages 3672--3681. PMLR, 2019.

\bibitem[Lei et~al.(2020)Lei, Yu, Berg, and Bansal]{lei2020tvr}
Jie Lei, Licheng Yu, Tamara~L Berg, and Mohit Bansal.
\newblock Tvr: A large-scale dataset for video-subtitle moment retrieval.
\newblock In \emph{Computer Vision--ECCV 2020: 16th European Conference, Glasgow, UK, August 23--28, 2020, Proceedings, Part XXI 16}, pages 447--463. Springer, 2020.

\bibitem[Leng et~al.(2024)Leng, Wu, Tan, Liu, Gan, Chen, and Gao]{dsrl}
Jiaxu Leng, Zhanjie Wu, Mingpi Tan, Yiran Liu, Ji Gan, Haosheng Chen, and Xinbo Gao.
\newblock Beyond euclidean: Dual-space representation learning for weakly supervised video violence detection.
\newblock In \emph{The Thirty-eighth Annual Conference on Neural Information Processing Systems}, 2024.

\bibitem[Lensink et~al.(2022)Lensink, Peters, and Haber]{net5}
Keegan Lensink, Bas Peters, and Eldad Haber.
\newblock Fully hyperbolic convolutional neural networks.
\newblock \emph{Research in the Mathematical Sciences}, 9\penalty0 (4):\penalty0 60, 2022.

\bibitem[Li et~al.(2024)Li, Chen, Xu, and Hu]{anomalydetection}
Huimin Li, Zhentao Chen, Yunhao Xu, and Junlin Hu.
\newblock Hyperbolic anomaly detection.
\newblock In \emph{Proceedings of the IEEE/CVF Conference on Computer Vision and Pattern Recognition}, pages 17511--17520, 2024.

\bibitem[Li et~al.(2019)Li, Xu, Yang, Chen, and Dong]{li2019w2vv++}
Xirong Li, Chaoxi Xu, Gang Yang, Zhineng Chen, and Jianfeng Dong.
\newblock W2vv++ fully deep learning for ad-hoc video search.
\newblock In \emph{Proceedings of the 27th ACM international conference on multimedia}, pages 1786--1794, 2019.

\bibitem[Liu et~al.(2025)Liu, Gao, Bai, Li, Shan, Dai, and Xia]{liu2025protecting}
Haitong Liu, Kuofeng Gao, Yang Bai, Jinmin Li, Jinxiao Shan, Tao Dai, and Shu-Tao Xia.
\newblock Protecting your video content: Disrupting automated video-based llm annotations.
\newblock In \emph{Proceedings of the Computer Vision and Pattern Recognition Conference}, pages 24056--24065, 2025.

\bibitem[Liu et~al.(2022)Liu, Liao, Wang, Wu, Li, Xia, and Xu]{liu2022multi}
Peidong Liu, Dongliang Liao, Jinpeng Wang, Yangxin Wu, Gongfu Li, Shu-Tao Xia, and Jin Xu.
\newblock Multi-task ranking with user behaviors for text-video search.
\newblock In \emph{Companion Proceedings of the Web Conference 2022}, pages 126--130, 2022.

\bibitem[Liu et~al.(2019{\natexlab{a}})Liu, Albanie, Nagrani, and Zisserman]{liu2019use}
Yang Liu, Samuel Albanie, Arsha Nagrani, and Andrew Zisserman.
\newblock Use what you have: Video retrieval using representations from collaborative experts.
\newblock \emph{arXiv preprint arXiv:1907.13487}, 2019{\natexlab{a}}.

\bibitem[Liu et~al.(2019{\natexlab{b}})Liu, Ott, Goyal, Du, Joshi, Chen, Levy, Lewis, Zettlemoyer, and Stoyanov]{liu2019roberta}
Yinhan Liu, Myle Ott, Naman Goyal, Jingfei Du, Mandar Joshi, Danqi Chen, Omer Levy, Mike Lewis, Luke Zettlemoyer, and Veselin Stoyanov.
\newblock Roberta: A robustly optimized bert pretraining approach.
\newblock \emph{arXiv preprint arXiv:1907.11692}, 2019{\natexlab{b}}.

\bibitem[Long et~al.(2020)Long, Mettes, Shen, and Snoek]{long2020searching}
Teng Long, Pascal Mettes, Heng~Tao Shen, and Cees~GM Snoek.
\newblock Searching for actions on the hyperbole.
\newblock In \emph{Proceedings of the IEEE/CVF conference on computer vision and pattern recognition}, pages 1141--1150, 2020.

\bibitem[Luo et~al.(2022)Luo, Ji, Zhong, Chen, Lei, Duan, and Li]{luo2022clip4clip}
Huaishao Luo, Lei Ji, Ming Zhong, Yang Chen, Wen Lei, Nan Duan, and Tianrui Li.
\newblock Clip4clip: An empirical study of clip for end to end video clip retrieval and captioning.
\newblock \emph{Neurocomputing}, 508:\penalty0 293--304, 2022.

\bibitem[McInnes et~al.(2018)McInnes, Healy, and Melville]{umap}
Leland McInnes, John Healy, and James Melville.
\newblock Umap: Uniform manifold approximation and projection for dimension reduction.
\newblock \emph{arXiv preprint arXiv:1802.03426}, 2018.

\bibitem[Meng et~al.(2025)Meng, He, Wang, Dai, Zhang, Zhu, Li, Wang, Zhang, and Jiang]{meng2025evdclip}
Guanghao Meng, Sunan He, Jinpeng Wang, Tao Dai, Letian Zhang, Jieming Zhu, Qing Li, Gang Wang, Rui Zhang, and Yong Jiang.
\newblock Evdclip: Improving vision-language retrieval with entity visual descriptions from large language models.
\newblock In \emph{Proceedings of the AAAI Conference on Artificial Intelligence}, pages 6126--6134, 2025.

\bibitem[Miech et~al.(2019)Miech, Zhukov, Alayrac, Tapaswi, Laptev, and Sivic]{miech2019howto100m}
Antoine Miech, Dimitri Zhukov, Jean-Baptiste Alayrac, Makarand Tapaswi, Ivan Laptev, and Josef Sivic.
\newblock Howto100m: Learning a text-video embedding by watching hundred million narrated video clips.
\newblock In \emph{Proceedings of the IEEE/CVF International Conference on Computer Vision}, pages 2630--2640, 2019.

\bibitem[Mun et~al.(2020)Mun, Cho, and Han]{mun2020local}
Jonghwan Mun, Minsu Cho, and Bohyung Han.
\newblock Local-global video-text interactions for temporal grounding.
\newblock In \emph{Proceedings of the IEEE/CVF Conference on Computer Vision and Pattern Recognition}, pages 10810--10819, 2020.

\bibitem[Nickel and Kiela(2017)]{poincareembedding}
Maximillian Nickel and Douwe Kiela.
\newblock Poincar\'{e} embeddings for learning hierarchical representations.
\newblock In \emph{Advances in Neural Information Processing Systems}. Curran Associates, Inc., 2017.

\bibitem[Nickel and Kiela(2018)]{lorentzraw}
Maximillian Nickel and Douwe Kiela.
\newblock Learning continuous hierarchies in the {L}orentz model of hyperbolic geometry.
\newblock In \emph{Proceedings of the 35th International Conference on Machine Learning}, pages 3779--3788. PMLR, 2018.

\bibitem[Pal et~al.(2025)Pal, van Spengler, di~Melendugno, Flaborea, Galasso, and Mettes]{hycoclip}
Avik Pal, Max van Spengler, Guido Maria~D'Amely di Melendugno, Alessandro Flaborea, Fabio Galasso, and Pascal Mettes.
\newblock Compositional entailment learning for hyperbolic vision-language models.
\newblock In \emph{The Thirteenth International Conference on Learning Representations}, 2025.

\bibitem[Peng et~al.(2023)Peng, Wen, Luo, Zhou, Yu, Wang, and Wu]{hypvd}
Xiaogang Peng, Hao Wen, Yikai Luo, Xiao Zhou, Keyang Yu, Yigang Wang, and Zizhao Wu.
\newblock Learning weakly supervised audio-visual violence detection in hyperbolic space, 2023.

\bibitem[Radford et~al.(2021)Radford, Kim, Hallacy, Ramesh, Goh, Agarwal, Sastry, Askell, Mishkin, Clark, et~al.]{CLIP}
Alec Radford, Jong~Wook Kim, Chris Hallacy, Aditya Ramesh, Gabriel Goh, Sandhini Agarwal, Girish Sastry, Amanda Askell, Pamela Mishkin, Jack Clark, et~al.
\newblock Learning transferable visual models from natural language supervision.
\newblock In \emph{International conference on machine learning}, pages 8748--8763. PMLR, 2021.

\bibitem[Shi et~al.(2024)Shi, Wen, Ji, Yang, Gao, and Zimmermann]{shi2024hover}
Ruiqi Shi, Jun Wen, Wei Ji, Menglin Yang, Difei Gao, and Roger Zimmermann.
\newblock {HOVER}: Hyperbolic video-text retrieval, 2024.

\bibitem[Song et~al.(2021)Song, Chen, Wu, and Jiang]{song2021spatial}
Xue Song, Jingjing Chen, Zuxuan Wu, and Yu-Gang Jiang.
\newblock Spatial-temporal graphs for cross-modal text2video retrieval.
\newblock \emph{IEEE Transactions on Multimedia}, 24:\penalty0 2914--2923, 2021.

\bibitem[Tan et~al.(2024)Tan, Lai, Zheng, and Hu]{tan2024siamese}
Chaolei Tan, Jianhuang Lai, Wei-Shi Zheng, and Jian-Fang Hu.
\newblock Siamese learning with joint alignment and regression for weakly-supervised video paragraph grounding.
\newblock In \emph{Proceedings of the IEEE/CVF Conference on Computer Vision and Pattern Recognition}, pages 13569--13580, 2024.

\bibitem[Tang et~al.(2025{\natexlab{a}})Tang, Wang, Peng, Meng, Luo, Chen, Chen, Wang, and Xia]{ACL25_CoPE}
Haomiao Tang, Jinpeng Wang, Yuang Peng, GuangHao Meng, Ruisheng Luo, Bin Chen, Long Chen, Yaowei Wang, and Shu-Tao Xia.
\newblock Modeling uncertainty in composed image retrieval via probabilistic embeddings.
\newblock In \emph{Proceedings of the 63rd Annual Meeting of the Association for Computational Linguistics (Volume 1: Long Papers)}, pages 1210--1222, 2025{\natexlab{a}}.

\bibitem[Tang et~al.(2025{\natexlab{b}})Tang, Yu, Gai, Zhuang, Xiong, Gou, and Wu]{tang2025missing}
Yuanmin Tang, Jing Yu, Keke Gai, Jiamin Zhuang, Gang Xiong, Gaopeng Gou, and Qi Wu.
\newblock Missing target-relevant information prediction with world model for accurate zero-shot composed image retrieval.
\newblock In \emph{Proceedings of the Computer Vision and Pattern Recognition Conference}, pages 24785--24795, 2025{\natexlab{b}}.

\bibitem[Van~Spengler et~al.(2023)Van~Spengler, Berkhout, and Mettes]{net3}
Max Van~Spengler, Erwin Berkhout, and Pascal Mettes.
\newblock Poincar{\'e} resnet.
\newblock In \emph{Proceedings of the IEEE/CVF International Conference on Computer Vision}, pages 5419--5428, 2023.

\bibitem[Vaswani et~al.(2017)Vaswani, Shazeer, Parmar, Uszkoreit, Jones, Gomez, Kaiser, and Polosukhin]{vaswani2017attention}
Ashish Vaswani, Noam Shazeer, Niki Parmar, Jakob Uszkoreit, Llion Jones, Aidan~N Gomez, {\L}ukasz Kaiser, and Illia Polosukhin.
\newblock Attention is all you need.
\newblock \emph{Advances in neural information processing systems}, 30, 2017.

\bibitem[Wang et~al.(2022)Wang, Chen, Liao, Zeng, Li, Xia, and Xu]{wang2022hybrid}
Jinpeng Wang, Bin Chen, Dongliang Liao, Ziyun Zeng, Gongfu Li, Shu-Tao Xia, and Jin Xu.
\newblock Hybrid contrastive quantization for efficient cross-view video retrieval.
\newblock In \emph{Proceedings of the ACM Web Conference 2022}, pages 3020--3030, 2022.

\bibitem[Wang et~al.(2024{\natexlab{a}})Wang, Zeng, Chen, Wang, Liao, Li, Wang, and Xia]{wang2024hugs}
Jinpeng Wang, Ziyun Zeng, Bin Chen, Yuting Wang, Dongliang Liao, Gongfu Li, Yiru Wang, and Shu-Tao Xia.
\newblock Hugs bring double benefits: Unsupervised cross-modal hashing with multi-granularity aligned transformers.
\newblock \emph{International Journal of Computer Vision}, 132\penalty0 (8):\penalty0 2765--2797, 2024{\natexlab{a}}.

\bibitem[Wang et~al.(2024{\natexlab{b}})Wang, Wang, Chen, Dai, Luo, and Xia]{gmmformerv2}
Yuting Wang, Jinpeng Wang, Bin Chen, Tao Dai, Ruisheng Luo, and Shu-Tao Xia.
\newblock Gmmformer v2: An uncertainty-aware framework for partially relevant video retrieval, 2024{\natexlab{b}}.

\bibitem[Wang et~al.(2024{\natexlab{c}})Wang, Wang, Chen, Zeng, and Xia]{GMMFORMER}
Yuting Wang, Jinpeng Wang, Bin Chen, Ziyun Zeng, and Shu-Tao Xia.
\newblock Gmmformer: Gaussian-mixture-model based transformer for efficient partially relevant video retrieval.
\newblock In \emph{Proceedings of the AAAI Conference on Artificial Intelligence}, 2024{\natexlab{c}}.

\bibitem[Wang et~al.(2025)Wang, Huang, Chen, Shi, Wan, Qiao, Yang, Wang, Li, and Ye]{wang2025empirical}
Zhihao Wang, Wenke Huang, Tian Chen, Zekun Shi, Guancheng Wan, Yu Qiao, Bin Yang, Jian Wang, Bing Li, and Mang Ye.
\newblock An empirical study of federated prompt learning for vision language model.
\newblock \emph{arXiv preprint arXiv:2505.23024}, 2025.

\bibitem[Wu et~al.(2023)Wu, Luo, Fang, Wang, and Ouyang]{wu2023cap4video}
Wenhao Wu, Haipeng Luo, Bo Fang, Jingdong Wang, and Wanli Ouyang.
\newblock Cap4video: What can auxiliary captions do for text-video retrieval?
\newblock In \emph{Proceedings of the IEEE/CVF Conference on Computer Vision and Pattern Recognition}, pages 10704--10713, 2023.

\bibitem[Yin et~al.(2024)Yin, Zhao, Wang, Xu, and Chen]{bgmnet}
Shukang Yin, Sirui Zhao, Hao Wang, Tong Xu, and Enhong Chen.
\newblock Exploiting instance-level relationships in weakly supervised text-to-video retrieval.
\newblock \emph{ACM Trans. Multim. Comput. Commun. Appl.}, 20\penalty0 (10):\penalty0 316:1--316:21, 2024.

\bibitem[Yu et~al.(2022)Yu, Nguyen, Gal, Ju, Chandra, Zhang, Bonnington, Mar, Wang, and Ge]{skin}
Zhen Yu, Toan Nguyen, Yaniv Gal, Lie Ju, Shekhar~S Chandra, Lei Zhang, Paul Bonnington, Victoria Mar, Zhiyong Wang, and Zongyuan Ge.
\newblock Skin lesion recognition with class-hierarchy regularized hyperbolic embeddings.
\newblock In \emph{International conference on medical image computing and computer-assisted intervention}, pages 594--603. Springer, 2022.

\bibitem[Zhang et~al.(2020)Zhang, Hu, Lee, Zhao, Chammas, Jain, Ie, and Sha]{zhang2020hierarchical}
Bowen Zhang, Hexiang Hu, Joonseok Lee, Ming Zhao, Sheide Chammas, Vihan Jain, Eugene Ie, and Fei Sha.
\newblock A hierarchical multi-modal encoder for moment localization in video corpus.
\newblock \emph{arXiv preprint arXiv:2011.09046}, 2020.

\bibitem[Zhang et~al.(2021)Zhang, Sun, Jing, Nan, Zhen, Zhou, and Goh]{zhang2021video}
Hao Zhang, Aixin Sun, Wei Jing, Guoshun Nan, Liangli Zhen, Joey~Tianyi Zhou, and Rick Siow~Mong Goh.
\newblock Video corpus moment retrieval with contrastive learning.
\newblock In \emph{Proceedings of the 44th International ACM SIGIR Conference on Research and Development in Information Retrieval}, pages 685--695, 2021.

\bibitem[Zhao et~al.(2023)Zhao, Wang, Liao, Wang, Duan, and Zhou]{zhao2023keyword}
Minyi Zhao, Jinpeng Wang, Dongliang Liao, Yiru Wang, Huanzhong Duan, and Shuigeng Zhou.
\newblock Keyword-based diverse image retrieval by semantics-aware contrastive learning and transformer.
\newblock In \emph{Proceedings of the 46th International ACM SIGIR Conference on Research and Development in Information Retrieval}, pages 1262--1272, 2023.

\end{thebibliography}
